\documentclass{article}
\usepackage{ijcai22}

\usepackage{times}
\usepackage{soul}
\usepackage{url}
\usepackage[breaklinks,hidelinks]{hyperref}
\usepackage[utf8]{inputenc}
\usepackage[small]{caption}
\usepackage{subcaption}
\usepackage{graphicx}
\usepackage{amsmath}
\usepackage{amsthm}
\usepackage{amssymb}
\usepackage{booktabs}
\usepackage[ruled]{algorithm2e}
\usepackage{algpseudocode}
\usepackage{fancyhdr}

\usepackage[textsize=tiny, textwidth=0.605in, colorinlistoftodos, disable]{todonotes}

\usepackage{newfloat}
\usepackage{listings}
\title{Offline Vehicle Routing Problem with Online Bookings:\\A Novel Problem Formulation with Applications to Paratransit}

\author{
Amutheezan Sivagnanam$^1$\and
Salah Uddin Kadir$^1$\and
Ayan Mukhopadhyay$^2$\and
Philip Pugliese$^3$\and\\
Abhishek Dubey$^2$\and
Samitha Samaranayake$^4$\And
Aron Laszka$^1$\\
\affiliations
$^1$University of Houston\and
$^2$Vanderbilt University\and\\
$^3$Chattanooga Area Regional Transportation Authority\and
$^4$Cornell University
}

\usepackage{pgfplots}
\pgfplotsset{compat=1.15}
\usepackage{pgfplotstable}
\usepgfplotslibrary{statistics}
\usepgfplotslibrary{patchplots}
\usepackage{csvsimple}

\usepackage{multicol}
\usepackage{multirow}
\usepackage[inline]{enumitem}
\usepackage[normalem]{ulem}

\usepackage[capitalise,noabbrev]{cleveref}

\pgfplotsset{
    select coords between index/.style 2 args={
        x filter/.code={
            \ifnum\coordindex<#1\fi
            \ifnum\coordindex>#2\fi
        }
    },
    rshift/.style={
        xshift=\pgfkeysvalueof{/pgfplots/rshift scale}
    },
    lshift/.style={
        xshift=-\pgfkeysvalueof{/pgfplots/lshift scale}
    },
    rshift2/.style={
        xshift=\pgfkeysvalueof{/pgfplots/rshift2 scale}
    },
    lshift2/.style={
        xshift=-\pgfkeysvalueof{/pgfplots/lshift2 scale}
    },
    rshift2 scale/.initial=2em,
    rshift scale/.initial=1em,
    lshift scale/.initial=1em,
    lshift2 scale/.initial=2em,
}

\definecolor{ColorCustomBlue}{rgb}{0,0,0.8}
\definecolor{ColorCustomRed}{rgb}{0.8,0,0}
\colorlet{ColorLegendPurple}{purple!90}
\colorlet{ColorLegendBrown}{brown!70}
\colorlet{ColorLegendBlue}{ColorCustomBlue!50}
\colorlet{ColorLegendRed}{ColorCustomRed!50}
\colorlet{ColorLegendGreen}{green!80!black}
\definecolor{ColorCustomGreen}{rgb}{0,0.3,0}
\pgfplotsset{
compat=1.16,
SmallBarPlot/.style={
    font=\footnotesize,
    ybar,
    width=\linewidth,
    ymin=0,
    xtick=data,
    xticklabel style={text width=0.8cm, align=center},
    xtick pos=left
},
BlueBars/.style={
    fill=blue!40, bar width=0.16
},
RedBars/.style={
    fill=red!40, bar width=0.16
},
GreenBars/.style={
    fill=green!75!black, bar width=0.16
}
}

\pgfplotsset{select coords between index/.style 2 args={
    x filter/.code={
        \ifnum\coordindex<#1\fi
        \ifnum\coordindex>#2\fi
    }
}}

\pgfplotsset{
    ylabel right/.style={
        after end axis/.append code={
            \node [rotate=90, anchor=north] at (rel axis cs:1,0.5) {#1};
        }   
    }
}

\newcommand{\Aron}[1]{\todo[backgroundcolor=yellow!10, linecolor=yellow!70!black]{\textbf{Aron:} #1}}

\newcommand{\ad}[1]{\todo[backgroundcolor=green!25]{\textbf{AD:} #1}}

\newcommand{\argmin}[0]{\ensuremath{\operatorname{argmin}}}

\newcommand{\vect}[1]{\ensuremath{\boldsymbol{#1}}}

\newcommand{\cut}[1]{}

\newif\ifExtendedVersion
\ExtendedVersiontrue

\begin{document}

\ifExtendedVersion
\pagestyle{fancy}
\renewcommand{\headrulewidth}{0pt}
\lhead{Accepted for publication in the proceedings of the 31st International Joint Conference on Artificial Intelligence (IJCAI 2022).}
\rhead{}
\cfoot{\thepage}
\setlength{\headsep}{2.5em}
\setlength{\footskip}{3.5em}
\setlength{\headheight}{0em}
\fi

\maketitle

\begin{abstract}%
Vehicle routing problems (VRPs) can be divided into two major categories:
offline VRPs, which consider a given set of trip requests to be served,
and online VRPs, which consider requests as they arrive in real-time.
Based on discussions with public transit agencies, we identify a real-world problem that is not addressed by existing formulations: booking trips with flexible pickup windows (e.g., 3 hours) in advance (e.g., the day before) and confirming tight pickup windows (e.g., 30 minutes) at the time of booking. 
Such a service model is often required in paratransit service settings, where passengers typically book trips for the next day over the phone.
To address this gap between offline and online problems, we introduce a novel formulation, the \emph{offline vehicle routing problem with online bookings}.
This problem is very challenging computationally since it faces the complexity of considering large sets of requests---similar to offline VRPs---but must abide by strict constraints on running time---similar to online VRPs.
To solve this problem, we propose a novel computational approach, which
combines an anytime algorithm with a learning-based policy for real-time decisions.
Based on a paratransit dataset obtained from the public transit agency of Chattanooga, TN, 
we demonstrate that our novel formulation and computational approach lead to significantly better outcomes in this setting than existing algorithms.%
\end{abstract}

\section{Introduction}

Vehicle routing problems (VRPs) can be divided into two major categories.
Offline VRPs consider a set of requests at once 
and optimize their assignment to planned vehicle routes
\cite{golden2008vehicle,laporte1992vehicle}. 
Online VRPs, on the other hand,
process requests as they arrive in real-time---either one-by-one or in small batches---and optimize their assignment to vehicle routes that may already be in progress \cite{toth2002vehicle,pillac2013review}.  
While online VRPs typically optimize fewer requests at a time, they are subject to stricter constraints on running time due to the online nature of the problems.

A socially beneficial application of VRPs is optimizing \emph{paratransit} services \cite{lave2000state}, which are curb-to-curb transportation services provided by public transit agencies for passengers who are unable to use fixed-route transit (e.g., passengers with disabilities). 
These services are crucial for providing transit accessibility to disadvantaged populations. 
Paratransit trips are typically booked at least one day in advance, which enables transit agencies to optimize routes as an offline VRP: before each day, an agency can optimize paratransit routes for that day based on all the requested pickup and drop-off locations and pickup time~\mbox{windows}.

However, based on discussions with public transit agencies, we identified a problem that is not addressed by existing VRP formulations. 
When passengers book trips over the phone, they often request \emph{broad pickup windows} (e.g., going for groceries in the afternoon, sometime between 2pm and 5pm).
While passengers may have no preference between pickup times within these broad windows, 
they do strongly prefer to know in advance when they will be picked up.
So, transit agencies must confirm a \emph{tight pickup window} (e.g., 30 minute interval within the broad window) at the time of booking.
The reason for this is very practical: vehicles may arrive at any time within the confirmed windows, and passengers need to be ready to be picked up.
This presents an interesting online optimization problem: \emph{how to select tight pickup windows at the time of booking, based on information available at the time, assuming that vehicle routes will be optimized as an offline VRP once all the trips have been booked?}

We formulate this as the \emph{offline vehicle routing problem with online bookings}.
We assume that trip requests with broad pickup windows are received one-by-one, and for each request, a tight pickup window must be selected in a matter of seconds.
At the end of the booking process, vehicle routes are optimized as an offline VRP based on the selected tight windows.
The objective of optimizing the tight pickup windows is to minimize the cost of the resulting offline VRP.
Note that this booking problem \emph{can be defined with respect to a wide range of offline VRP formulations} (that consider pickup windows), so our framework could be applied to a range of real-world problems where tight pickup windows must be chosen during booking (e.g., scheduling the delivery of refrigerated goods or dial-a-ride services).
In this paper, we consider an offline VRP formulation that models paratransit~services. 

This problem is very challenging computationally since it faces the complexity of considering large sets of trips---similar to offline VRPs---but must abide by strict limits on running time---similar to online VRPs.
To address this challenge, we propose a novel computational approach that combines an \emph{anytime algorithm} with a \emph{reinforcement-learning based policy}.\cut{ The anytime algorithm is executed between consecutive requests and provides supporting input for the policy, taking advantage of the extra time between requests (e.g., there may be 5 minutes between consecutive calls, but only a few seconds to respond during a call).
The policy is responsible for selecting tight pickup windows, taking advantage of the input received from the anytime algorithm.
The advantage of a learning based policy is that its computational cost is very low once it has been trained on historical data.}
We demonstrate that our novel formulation and computational approach lead to significantly better outcomes in the paratransit-booking setting than existing algorithms using real-world data from a public transit agency.




\newcommand{\VRP}[0]{\ensuremath{\textit{VRP}}}
\newcommand{\RequestsVector}[0]{\ensuremath{\langle T_i \rangle}}
\newcommand{\RequestsSubVector}[1]{\ensuremath{\langle T_j \rangle}_{j=1,\ldots,#1}}
\newcommand{\WindowsVector}[0]{\ensuremath{\langle w_i \rangle}}
\newcommand{\WindowsSubVector}[1]{\ensuremath{\langle w_i \rangle}_{j=1,\ldots,#1}}
\newcommand{\BroadWindowsVector}[0]{\ensuremath{\langle W_i \rangle}}
\newcommand{\LocationVector}[0]{\ensuremath{\langle L_i \rangle}}
\newcommand{\RunsVector}[0]{\ensuremath{\langle R_i \rangle}}
\renewcommand{\RequestsVector}[0]{\ensuremath{\vect{T}}}
\renewcommand{\RequestsSubVector}[1]{\ensuremath{\langle T_1, \ldots, T_{#1} \rangle}}
\renewcommand{\WindowsVector}[0]{\ensuremath{\vect{w}}}
\renewcommand{\WindowsSubVector}[1]{\ensuremath{\langle w_1, \ldots, w_{#1} \rangle}}
\renewcommand{\BroadWindowsVector}[0]{\ensuremath{\vect{W}}}
\renewcommand{\LocationVector}[0]{\ensuremath{\vect{L}}}
\renewcommand{\RunsVector}[0]{\ensuremath{\vect{R}}}
\newcommand{\Feasible}[0]{\ensuremath{R}}
\newcommand{\Cost}[0]{\ensuremath{C}}
\renewcommand{\Feasible}[0]{\ensuremath{\mathcal{R}}}
\renewcommand{\Cost}[0]{\ensuremath{\mathcal{C}}}

\section{Model and Problem Formulation}
\label{sec:model}

We formulate the \emph{offline VRP with online bookings} by first introducing an \emph{offline vehicle routing problem}, which models the optimization of allocating trip requests to vehicle routes once all the trip requests have been booked and their tight pickup windows have been confirmed.
Building on this offline problem, we then formulate the \emph{online booking problem}, which models the optimization of tight pickup windows in real time, assuming that vehicle routes will be optimized afterwards.
\ifExtendedVersion%
\cref{table:symbol_list} in \cref{app:symbols} provides a list of symbols. 
\else%
Table~1 in Appendix~A provides a list of symbols~\cite{sivagnanam2022offline}. 
\fi

\subsection{Vehicle Routing Problem with Time Windows}
\label{sec:offline_VRP_short}

Offline VRPs with time windows
is a family of classical combinatorial problems.
Here, we introduce the offline VRP formulation that we employ in our experiments, which we developed to model paratransit services.
However, it is important to note that the online bookings problem could be defined with respect to a wide range of offline VRP formulations with pickup time windows, and our proposed solution approach can incorporate existing offline VRP solvers for these problems.
Since our offline VRP is a minor variation of classical formulations, here we provide only a concise summary of this problem, which is sufficient for formulating the novel online bookings problem.
\ifExtendedVersion%
Due to lack of space, we provide a detailed formal definition in~\cref{app:offline_problem}.
\else%
Due to lack of space, we provide a full formal definition in Appendix~B~\cite{sivagnanam2022offline}.
\fi


\paragraph{Problem Input}
The input of the offline VRP problem is
an ordered set of \emph{trip requests} $\RequestsVector = \langle T_1, T_2, \ldots, T_n\rangle$, where each trip request $T_i$ contains a pickup location $L^{\textit{pickup}}_i$, a drop-off location $L^{\textit{dropoff}}_i$, and the number of passengers to be transported $P_i$;
and 
a corresponding ordered set of \emph{tight pickup time windows} $\WindowsVector = \langle w_1, w_2, \ldots, w_n\rangle$, where each time window $w_i$ is defined by an earliest $w_i^\textit{start}$ and latest $w_i^\textit{end}$ pickup time. 
The input also includes
constants, such as the maximum allowed duration $D^\textit{maxroute}$ of a vehicle route, 
the passenger capacity $V$ of the vehicles, and so on (see \ifExtendedVersion\cref{app:offline_problem}\else{}Appendix~B\fi). 
For ease of presentation, we will not list these constants explicitly and represent a VRP instance simply as $(\RequestsVector, \WindowsVector)$, assuming that the constants are provided implicitly. 

\paragraph{Solution and Objective}
A solution to the offline VRP problem is
a set of \emph{vehicle routes} $\vect{R} = \{ R_1, R_2, \ldots, R_m\}$, where each route is an ordered set of pickups $L^{\textit{pickup}}_i$ and drop-offs $L^{\textit{dropoff}}_i$ (note that trips may be combined in a route, i.e., pickups and drop-offs of different trips may be interleaved).
A set of vehicle routes $\vect{R}$ is a feasible solution if {each pickup $L^{\textit{pickup}}_i$ is included in exactly one route $R_j$, the corresponding dropoff $L^{\textit{dropoff}}_i$ is also included in $R_j$}, and every route satisfies time constraints (
passengers are picked up within the pickup time windows, 
travel times between locations are respected,%
\cut{route durations do not exceed the maximum allowed,}
etc.)
and vehicle capacity constraints (see \ifExtendedVersion\cref{app:offline_problem}\else{}Appendix~B\fi).
\cut{The passengers of trip request $T_i$ must be dropped off by $w_i^\text{end} + D^{\textit{travel}}_{L^{\textit{pickup}}_i,L^{\textit{dropoff}}_i}$ at latest, where $dw^\textit{end}_i$ is the latest pickup time and $D^{\textit{travel}}_{L^{\textit{pickup}}_i,L^{\textit{dropoff}}_i}$ is the direct travel time (i.e., without any detours) from the pickup location $L^{\textit{pickup}}_i$ to the dropoff location $L^{\textit{dropoff}}_i$.}
We let $\Feasible(\RequestsVector, \WindowsVector)$ denote the set of feasible solutions for a VRP instance $(\RequestsVector, \WindowsVector)$.

The cost of a solution depends on the number of vehicle routes and the duration of each route (see \ifExtendedVersion\cref{app:offline_problem}\else{}Appendix~B\fi). 
By letting $\Cost(\vect{R})$ denote the cost of a solution $\vect{R}$,
we can express the offline VRP problem as $\argmin_{\vect{R} \in \Feasible(\RequestsVector, \WindowsVector)} \Cost(\vect{R})$.
Finally, we let $\VRP^*(\RequestsVector, \WindowsVector)$ denote the total cost of an optimal solution for problem instance $(\RequestsVector, \WindowsVector)$. 
That is,
$\VRP^*(\RequestsVector, \WindowsVector) = \min_{\vect{R} \,\in\, \Feasible(\RequestsVector, \WindowsVector)} \Cost(\vect{R})$.
Since there is a vast literature on solving offline VRPs, we assume that an offline VRP solver (i.e., heuristic or approximation algorithm for $\VRP^*$) is given, and focus on the online bookings problem in this paper.

\subsection{Online Bookings Problem}

Building on the offline VRP formulation, we now introduce the online bookings problem.
In this real-time decision problem, trip requests $T_1, T_2, \ldots$ are received one-by-one, and 
each trip request $T_i$ is accompanied by a \emph{broad pickup window} $W_i$.
Our goal is to select a tight pickup window $w_i \subset W_i$ for each  trip request~$T_i$ in real-time (i.e., in a few seconds after the request is received),
so that once we have received all the trip requests $\RequestsVector$ and selected all the pickup windows~$\WindowsVector$, the total cost $\VRP^*(\RequestsVector, \WindowsVector)$ of the resulting offline VRP is minimized.
To model uncertainty and expectations about future requests, we assume that the sets of trip requests and broad pickup windows $(\RequestsVector, \BroadWindowsVector)$ are drawn at random from a known probability distribution $\mathcal{D}$ (note that the number of requests is variable).
So, each decision is
based on previously received requests \cut{information available at the time (i.e., received requests $\RequestsSubVector{i}$ and previously selected pickup windows $\WindowsSubVector{i-1}$) as well as}and expectation of future ones (i.e., distribution $\mathcal{D}$).

\paragraph{Problem Input}
Formally, the input for the $i$th decision is
the ordered set of trip requests (including the $i$th request) $\RequestsSubVector{i}$,
the ordered set of previously selected tight pickup windows (up to the $(i-1)$th request) $\WindowsSubVector{i-1}$,
and a broad pickup window $W_i$, which specifies the earliest $W_i^\textit{start}$ and latest $W_i^\textit{end}$ pickup time for request $T_i$.
The input also includes the probability distribution~$\mathcal{D}$,
the maximum duration of tight pickup windows $D^\textit{window}$ (e.g., 30 minutes),
and any additional inputs that are required by the offline VRP (e.g., vehicle capacity, maximum route duration); for ease of presentation, we will not list these additional inputs explicitly. 

\paragraph{Decision Space and Objective}
The output of the $i$th decision is a tight pickup window $w_i$ that is at most $D^\textit{window}$ long (i.e., $w_i^\textit{end} - w_i^\textit{start} \leq D^\textit{window}$) and falls within the broad window (i.e., $W_i^\textit{start} \leq w_i^\textit{start} \leq w_i^\textit{end} \leq W_i^\textit{end}$).

Whether a decision $w_i$ is optimal 
depends not only on the received requests and on our expectation of future requests, but also on how we will respond to those future requests.
Thus, instead of trying to formulate the online booking problem as optimizing each decision $w_i$, we formulate it as optimizing a decision-making policy $\mu$,  which maps each input $(\RequestsSubVector{i}, \WindowsSubVector{i-1}, W_i)$ to a tight pickup window~$w_i$.
Formally, our goal is to find an \emph{optimal decision policy}~$\mu^*$, which minimizes the expected cost of the resulting offline VRP instance $(\RequestsVector, \WindowsVector)$:
\begin{align*}
\argmin_{\mu} \mathbb{E}_{(\RequestsVector, \BroadWindowsVector) \sim \mathcal{D}} \bigg[ & \VRP^*\left(\RequestsVector, \WindowsVector \right) \nonumber\\[-0.5em]
& ~~~ \Big|_{w_i = \mu\left( \RequestsSubVector{i}, \WindowsSubVector{i-1}, W_i \right)}
\bigg] .
\end{align*}
Since the decision problem is subject to real-time constraints (i.e., when someone calls over the phone to book a paratransit trip, the transit agency must respond during the phone call, within seconds), we must be able to evaluate the policy~$\mu^*$ in a matter of seconds for any input.
\section{Solution Approach}
\label{sec:solution}

\subsection{Anytime Algorithm}

The online booking problem is computationally challenging since it incorporates the offline VRP into its objective\cut{  (tight pickup windows must be selected considering how the resulting offline VRP could be solved)}, which is a computationally-hard combinatorial optimization problem 
(\cite{lenstra1981complexity}).
Indeed, existing approaches for solving offline VRPs are not well suited for real-time applications (i.e., finding solutions within a matter of seconds). 

To address this challenge, we propose \emph{performing computation between consecutive decisions}.
While there is limited time for each real-time decision\cut{ (since extended periods of waiting would not be acceptable in practice for an over-the-phone booking)}, there is significantly more time between consecutive decisions (i.e., from when a tight window is selected to when the next request is received).
We can take advantage of this extra time 
by continuously working on a vehicle-routing solution, which can then be used as supporting input in the next real-time decision.

Unfortunately, the amount of time between consecutive requests is not known in advance since calls arrive at random times  (note that the arrival time of a request is different from its broad pickup window).
Thus, we propose to
employ an \emph{anytime VRP algorithm}, which we can start after each real-time decision and stop when the next request arrives.

\subsubsection{Anytime-supported Online Bookings Problem}

Based on the above ideas, we reformulate our online decision problem as the \emph{anytime-supported online bookings \mbox{problem}}.

\paragraph{Policy Input and Decision Space}
The input for the $i$th decision is the same as before, but now also includes a feasible VRP solution  $\vect{R}{}^{(i-1)}$ (i.e., a set of routes), provided by the $(i\!-\!1)$th execution of the anytime algorithm, which we specify below.
For ease of exposition, we define $\RunsVector{}^{(0)}=\emptyset$ for the very first request $T_1$.
The output of the $i$th decision is also the 
same as before, but now also includes a feasible VRP solution $\hat{\vect{R}}{}^{(i)}$ (i.e., $\hat{\vect{R}}{}^{(i)} \in \Feasible(\RequestsSubVector{i}, \WindowsSubVector{i})$, provided as supporting input for the $i$th execution of the anytime algorithm.
Note that we could omit the VRP solution $\hat{\vect{R}}{}^{(i)}$ from the output of the online decision and let the anytime algorithm assign the new request to a route. 
However, we found that providing a feasible solution as a starting point for the anytime algorithm is very beneficial in practice since the selection of the pickup window must consider anyway how the request  will ``fit'' into the routes.
Also note that finding a feasible VRP solution does not introduce a computational challenge since we can let the decision policy assign each new request to a new route and leave existing routes unchanged (achieving feasibility, but leaving all of the VRP optimization to the anytime algorithm); we of course train our policy to provide better~solutions.

\paragraph{Anytime Algorithm Input and Output}
The input for the $i$th execution of the anytime algorithm consists of
the trip requests $\RequestsSubVector{i}$, 
the tight pickup windows $\WindowsSubVector{i}$, and
a feasible VRP solution $\hat{\vect{R}}{}^{(i)}$, provided by the $i$th decision of the policy.
The output of the $i$th execution of the anytime algorithm is an improved feasible VRP solution $\vect{R}{}^{(i)}$, provided for the $(i\!+\!1)$th decision of the policy. 

The objective of the anytime algorithm $\alpha$ is to find a minimum-cost feasible solution for the VRP instance $\left(\RequestsSubVector{i}, \WindowsSubVector{i}\right)$, using the solution $\hat{\vect{R}}{}^{(i)}$ as a \emph{warm start} (e.g., as initial solution for simulated annealing): 
\begin{align*}
\vect{R}{}^{(i)} &= \alpha\big(\RequestsSubVector{i}, \WindowsSubVector{i}, \hat{\vect{R}}{}^{(i)}\big) \\ &\approx \argmin_{\vect{R} \,\in\, \Feasible(\RequestsSubVector{i}, \WindowsSubVector{i})} \Cost(\vect{R}) .
\end{align*}
Note that the objective of the anytime algorithm does not consider future requests, 
only ones that have been received. 
In our experiments, we found that we can attain very good performance by letting the decision policy handle expectations about future requests, and restricting the anytime algorithm to optimizing for requests that have been received.

\paragraph{Optimal Decision Policy}
Finally, we can reformulate our goal for the online bookings problem as finding an optimal decision policy $\mu^*$ for selecting tight pickup windows, supported by the anytime algorithm~$\alpha$:
\begin{align*}
&\argmin_{\mu}  \mathbb{E}_{(\RequestsVector, \BroadWindowsVector) \sim \mathcal{D}} \Big[ \VRP^*\left(\RequestsVector, \WindowsVector \right) \nonumber \\[-0.25em]
 & ~~~\Big|_{\big(w_i, \hat{\vect{R}}{}^{(i)}\!\big) = \mu\big( \RequestsSubVector{i}, \WindowsSubVector{i-1}, W_i, \vect{R}^{(i-1)} \big), \,\, \vect{R}^{(i)} = \alpha \big( \ldots \big)}
\Big] .
\end{align*}
Note that the anytime algorithm $\alpha$ can be implemented using an existing offline VRP solver---as long as it is anytime.
\cut{In \cref{sec:comp_workflow}, we will introduce an anytime simulated-annealing algorithm $\alpha^\textit{SimAnn}$ that is tailored to our problem setting and can utilize the supporting input~$\hat{\vect{R}}{}^{(i)}$.}
%

\subsection{Decision Policy}
\label{sec:decision_policy}

We can view online bookings as a Markov decision process (MDP)
: 
a decision input $(\RequestsSubVector{i},$ $\WindowsSubVector{i-1}, W_i, \vect{R}{}^{(i-1)})$ is a \emph{state of the environment}, 
a decision output $\big(w_i, \hat{\vect{R}}{}^{(i)}\big)$ is an \emph{action}, and running the anytime algorithm until a new request arrives at random is the \emph{state transition} (reaching a terminal state when no more requests arrive for the day).
To formulate an MDP, we also have to define the \emph{immediate cost} (i.e., immediate negative reward) that we incur for taking an action. 
The online bookings problem quantifies costs at the end of the day---after the last decision---based on the total cost of the resulting offline VRP instance~$(\RequestsVector, \WindowsVector)$.
Thus, the immediate cost $c_i$ incurred for the $i$th decision is 
\begin{equation*}
c_i = \begin{cases}
  0 & \text{ if } i < |\RequestsVector| \\
  \VRP^*(\RequestsVector, \WindowsVector) & \text{ if } i = |\RequestsVector| .
\end{cases}
\end{equation*}
By formulating the online bookings problem as an MDP, we enable the application of \emph{reinforcement learning} (RL) to find an optimal decision policy $\mu^*$.
The advantage of RL is that once a policy $\mu^*$ has been trained, the computational cost of execution is low, which is crucial for real-time decisions.

To find an optimal policy, an RL algorithm gathers experiences by repeatedly interacting with the environment in a number of training episodes, recording the experienced states, actions, and immediate costs. In \cut{the online bookings problem}our case, experiences can be gathered by running simulations of the online bookings process, where an input $(\vect{T}, \vect{w})$ is drawn at random from distribution $\mathcal{D}$ for each episode.
From these experiences, an RL algorithm can learn an optimal policy that minimizes the expected cumulative cost\cut{, often considering future costs with a temporal discount factor}.
Many popular RL algorithms, such as deep Q-learning (DQN) and its variants, learn a policy by learning 
an \emph{action-value function},
which estimates the expected cumulative cost when taking a given action in a given state (e.g., using the recursive Bellman equation to consider future costs).
Once the action-value function has been learned, the optimal policy is  to simply choose an action that minimizes the action-value function in the current state.

%

\paragraph{Cost Design}

RL\cut{Reinforcement learning} approaches
face two significant challenges in our environment.
First, the total cost of the offline VRP instance depends as much on the decision inputs (e.g., on the sheer number of trip requests) as it does on the decisions of the 
policy. 
Since the decision inputs\cut{ (i.e., trip requests and broad windows)} are random and vary significantly (e.g., there are significant differences between the number of trip requests each day), experiences will be extremely noisy and difficult to learn from.
Second, simulating the environment is very expensive computationally since each state transition\cut{ (i.e., progression to the next trip request)} requires running the anytime algorithm for a significant amount of time (e.g., 5 minutes of running time to obtain a single experience).
This greatly exacerbates the problem of noisy experiences since a low number of noisy experience can lead to very inaccurate action-value functions. 

To address these challenges, we replace the original immediate cost $c_i$ of the MDP with a \emph{shaped cost}~$\tilde{c}_i$, which assigns a cost to each individual decision: 
\begin{align*}
\tilde{c}_i = &\VRP^*(\RequestsVector, &&\langle w_1, \ldots, w_{i-1}, w_i, W_{i+1}, \ldots, W_{|\RequestsVector|}\rangle) \nonumber \\
& - \VRP^*(\RequestsVector, \!\!\!\!\!&&\langle w_1, \ldots, w_{i-1}, W_{i}, W_{i+1}, \ldots, W_{|\RequestsVector|} \rangle ) .
\end{align*}

The rationale behind the above formulation is to capture the impact of narrowing down the broad window $W_i$ to a tight window $w_i$ in the $i$th decision.
This shaped cost $\tilde{c}_i$ formulation has two advantages. First, notice that
\begin{equation*}
\sum_{i=1}^{|\RequestsVector|} \tilde{c}_i = \Bigg(\sum_{i=1}^{|\RequestsVector|} c_i \Bigg) - \VRP^*(\RequestsVector, \BroadWindowsVector) .
\end{equation*}
In other words, the difference between the original cumulative cost $\sum c_i$ and the shaped cumulative cost $\sum \tilde{c_i}$ is removing a part of the cost that does not depend on the decisions
(i.e., removing the cost $\VRP^*(\RequestsVector, \BroadWindowsVector)$ of a ``baseline'' VRP instance $(\RequestsVector, \BroadWindowsVector)$, where we could schedule pickups for any time within the broad windows $\BroadWindowsVector$).
So, shaped costs~$\tilde{c}_i$ capture only the impact of the decisions, thereby reducing~noise.

Second, since the shaped cost $\tilde{c}_i$ captures the impact of a decision considering all future requests (i.e., considering the complete ordered sets $\vect{T}$ and $\vect{W}$), we can use it to quantify the value of a decision without taking future costs into account.
In other words, the expected impact of a decision on the total cost $\VRP^*(\RequestsVector, \WindowsVector)$ is captured by the immediate shaped cost~$\tilde{c}_i$ since this cost~$\tilde{c}_i$ is the increase in the cost of the offline VRP with all the trip requests $\vect{T}$.
Hence, we can use experiences to learn a value function that estimates the expected immediate shaped cost~$\tilde{c}_i$ of a given action in a given state; 
once the value function has been learned, our policy is to simply choose an action that minimizes the value (i.e., cost) in the current state.
This significantly reduces the complexity of learning and, thus, the number of experiences required.


Note that during training, we can estimate shaped cost $\tilde{c}_i$  since we simulate the environment, so we can generate all the trip requests $(\vect{T}, \vect{W})$ before feeding them to the policy one-by-one.
Once the value function has been learned, calculating the shaped cost $\tilde{c}_i$ is no longer necessary since the policy is to choose an action that minimizes the learned value (i.e., cost) function in the current state.
Finally, note that calculating $\VRP^*$ is computationally hard, so we can use a heuristic VRP solver\cut{ (e.g., $\alpha^\textit{Greedy}$ from \cref{sec:comp_workflow})} during training to estimate shaped cost~$\tilde{c}_i$.

\paragraph{Value Function}

Since our value function considers only the immediate shaped cost $\tilde{c}_i$, we can apply a simplified version of the popular DQN algorithm. 
Before applying DQN, we have to discretize the action space, that is, constrain each decision to a discrete set of choices (e.g., pickup times must be multiples of 15 minutes).
Such discretization is natural for transit agencies that prefer ``round'' pickup times.

Our goal is to learn the value function $Q$:
\begin{equation*}
Q\big( \underbrace{\!\RequestsSubVector{i}, \WindowsSubVector{i-1}, W_i, \vect{R}{}^{(i-1)}\!}_{\textit{state}},  \underbrace{w_i, \hat{\vect{R}}{}^{(i)}\!}_{\textit{action}} \big) \approx  \tilde{c}_i .
\end{equation*}
Once we have learned the value function $Q$, our policy $\mu^*$ is to iterate over the actions and select one that minimizes the cost in the current state: 
\begin{equation*}
    \mu^*(\textit{state}) = \argmin_{w_i, \hat{\vect{R}}{}^{(i)}} Q(\textit{state}, w_i, \hat{\vect{R}}{}^{(i)}) .
\end{equation*}

To enable learning, we represent $Q$ as a neural network, which we initialize with random weights.
During training, we execute our policy in simulated environments, where the inputs $(\vect{T}, \vect{w})$ are drawn at random from distribution $\mathcal{D}$.
We collect experiences, that is, tuples of state, action, and immediate cost, and we use these experiences to train the neural network.
As is standard in RL, we also include random actions in the training to balance exploration and exploitation. 


\paragraph{Features}
Learning the value function $Q$ poses one last challenge due to the size and complexity of the state space (i.e., space of all possible decision inputs, including possible sets of requests $\RequestsSubVector{i}$ and feasible sets of runs $\vect{R}{}^{(i-1)}$).
While similar state spaces have been considered in prior work (e.g.,~\cite{joe2020deep,james2019online}),
the challenge in our problem is exacerbated by the prohibitively high computational cost of the environment (one state transition may require running an anytime algorithm for 5 minutes), which limits the number of experiences that we can~collect. 

To reduce the number of experiences required for training the value function, we map the large and complex space of states and actions to a low- and fixed-dimensional space of \emph{feature vectors}, and we replace the input of the value function $Q$ with a feature vector.
Specifically, we map each state-action pair to a vector of features, which consider how well the action fits the current state (i.e., \ifExtendedVersion{}how well it fits \else\fi 
the previously chosen tight windows $\WindowsSubVector{i-1}$ and  vehicle routes $\vect{R}{}^{(i-1)}$) and our expectation of future trip requests (i.e., distribution $\mathcal{D}$). 

Features that consider how well the action $(w_i, \hat{\vect{R}}{}^{(i)})$ fits the previously chosen tight windows $\WindowsSubVector{i-1}$ and vehicle routes $\RunsVector{}^{(i-1)}$\cut{ (i.e., latest solution from the anytime algorithm $\alpha$)} 
include (1) the increase in the duration of routes due to taking the action (compared between $\hat{\vect{R}}{}^{(i)}$ and $\RunsVector{}^{(i-1)}$), (2) the increase in the driving distance of routes, and (3) the ``tightness'' of the route schedule $\hat{\vect{R}}{}^{(i)}$, that is, how much time slack is left in the route before and after serving trip request $T_i$.
Features that consider  the distribution $\mathcal{D}$ include the expected number of trips requests whose pickup locations are nearby the pickup location $L_i^\textit{pickup}$ of request $T_i$ and/or whose drop-off locations are nearby the drop-off location~$L_i^\textit{dropoff}$ 
and/or whose broad pickup windows are around the same time as window $w_i$, as well as the expected number of future trip requests (i.e., expectation of $|\vect{T}| - i$).
For any state and action, these features can be calculated at a relatively low computational cost based on historical data, which enables real-time application. 
Due to lack of space, we provide a formal description of these features in 
\ifExtendedVersion\cref{app:features}\else{}Appendix~C \cite{sivagnanam2022offline}\fi.
\cut{One subset of the features is based on the distribution of trip requests $\mathcal{D}$.
These features consider the expected number of trip requests whose pickup locations are nearby the pickup location $L_i^\textit{pickup}$ of request $T_i$ and/or whose drop-off locations are nearby the drop-off location~$L_i^\textit{dropoff}$ 
and/or  whose broad pickup windows are around the same time as window $w_i$.
We include four features in the input of the value function $Q$: expected number of requests that match (1) all three criteria, (2) pickup location and pickup window criteria, (3) drop-off location and pickup window criteria, and (4) only the pickup window criterion.}%
\cut{We also include a feature that considers the expected number of future trip requests (i.e., expectation of $|\vect{T}| - i$).
In practice, we can easily calculate of all of these features based on historical data.}

%
%
%
%
%
%
\cut{Here, we summarize our computational approach by providing an overview of the training process, 
and we introduce an anytime simulated-annealing $\alpha^\textit{SimAnn}$ algorithm that is tailored to the paratransit VRP formulation.}
\paragraph{Training Process}
Before we begin training, we initialize the value function $Q$, which is represented by a neural network, with random weights.
We then train the policy $\mu$ (i.e., value function~$Q$) over a number of training episodes, where each episode is a simulation of the online booking process with a random input $(\RequestsVector, \BroadWindowsVector)$ drawn
from the probability distribution $\mathcal{D}$, which we estimate based on historical data. 
To simulate the online booking process, we process the trip requests~$\RequestsVector$ one-by-one, first applying the decision policy $\mu$ and then the anytime algorithm $\alpha$ for each request $T_i$.
To apply the policy~$\mu$, we calculate the feature vector for every action $(w_i, \hat{\vect{R}}{}^{(i)})$, evaluate the value function $Q$ over these feature vectors, and select the action that minimizes the value (i.e., cost).
Next, we run the anytime algorithm $\alpha$, which provides supporting input for the next policy decision.
We terminate the algorithm after a random amount of running time, which models the random inter-arrival time of requests (based on historical data).
Then, we repeat with next request $T_{i+1}$; or with the next episode if~$i = |\RequestsVector|$.

After each simulated decision, we collect an experience (i.e., a tuple of the feature vector and the shaped cost $c_i$) by calculating the shaped cost $\tilde{c}_i$ using $\RequestsVector$, $\BroadWindowsVector$, $\WindowsSubVector{i}$, and a heuristic for $\VRP^*$.
We use these experiences to train the value function $Q$ (i.e.,  the neural network).
In the beginning, the policy $\mu$ chooses actions at random since function~$Q$ is initialized randomly; but as we train function $Q$ using more and more experiences, the policy improves and converges to an optimum $\mu^*$ (given feature-vector inputs and objective $c_i$).
To balance exploration and exploitation and to avoid converging to a local optimum, we occasionally take random actions during training, as is standard in~RL.

\section{Evaluation}
\label{sec:numerical}

\definecolor{ColorCustomGreen}{rgb}{0, 0.8, 0}
\colorlet{ColorOur}{ColorCustomGreen} 
\colorlet{ColorLegendOur}{ColorOur!50}
\colorlet{ColorVroom}{red}
\colorlet{ColorLegendVroom}{ColorVroom!50}
\colorlet{ColorGoogle}{orange}
\colorlet{ColorLegendGoogle}{ColorGoogle!50}
\colorlet{ColorOurWOS}{blue}
\colorlet{ColorLegendOurWOS}{ColorOurWOS!50}

\subsection{Dataset and Experimental Setup}

\paragraph{Paratransit Data}
To evaluate our proposed approach, we obtained real-world paratransit data from~CARTA, the public transit agency of Chattanooga, TN, a mid-sized U.S. city.
This dataset spans 180 days of paratransit service, \cut{from the first half of 2021.}%
with an average of 140 trips per day (minimum of 7 and maximum of 234 trips). \cut{On average, the agency served 140 paratransit trips per day, with a minimum of 7 and a maximum of 234  trips per day.}
Each trip has an associated pickup and drop-off location (specified as latitude-longitude pairs), the number of passengers, and the scheduled pickup time.
The anonymized dataset and our software implementation are available at \url{https://github.com/smarttransit-ai/ijcai22}

Based on input from the agency, we instantiate our model with vehicle capacity $V = 9$, 
maximum route duration $D^\textit{maxroute} = 10$ hours, 
%
%
and maximum tight pickup window duration $D^\textit{window} = 30$ minutes. 
Since the agency did not record the requested broad windows (current over-the-phone booking process is manual), we assume each broad window~$W_i$ to be 3 hours long (which is a practical value for the service) and centered around the scheduled pickup time. 

\paragraph{Experimental Setup}
We calculate travel times between the locations using road-network data from OpenStreetMaps. The road network contains 10,788 nodes (i.e., intersection) and 28,100 edges (i.e., roads).
For calculating feature vectors, 
we consider two locations to be nearby if they have the same ZIP code. 
We assume that the transit agency operating paratransit services must serve all the passenger requests according to the Americans with Disabilities Act.  


%
We implemented our framework in Python~3.8.
To provide anytime algorithms, we implemented a heuristic greedy $\alpha^\textit{Greedy}$ and a meta-heuristic simulated annealing algorithm $\alpha^\textit{SimAnn}$, which we use in tandem as our anytime VRP solver $\alpha^\textit{SimAnn+Greedy}$.
Since these are based on standard techniques, we describe them in \ifExtendedVersion\cref{app:anytime_algorithms}\else{}Appendix~D~\cite{sivagnanam2022offline}\fi.
During both training and evaluation, we let the anytime algorithm run for 5 minutes on average.
In our experiments, we consider two offline VRP solvers: {VROOM\ifExtendedVersion~\cite{vroom}\fi} and the Google OR-Tools Vehicle Routing framework\ifExtendedVersion~\cite{googleortools}\fi. \cut{For Google OR-Tools, we developed a wrapper to provide support for our paratransit objective and constraints (e.g., route duration).}%
We do not impose a running time limit on either VROOM or Google OR-Tools. 
%
To represent the value function $Q$, we use a neural network with one input layer, one hidden layer (64 neurons, ReLU activation), and one output layer (linear activation). 
To train the network, we use the Adam optimizer\ifExtendedVersion~\cite{kingma2014adam}\fi~from the Keras library.



\subsection{Results}

We provide supplementary numerical results in \ifExtendedVersion\cref{app:numerical}\else{}Appendix~E \cite{sivagnanam2022offline}\fi.

\paragraph{Running Time}
We  run  all  algorithms
on\cut{ a  computer with}  an Intel  Xeon  E5-2680 28-core CPU\cut{,  which  has  28  cores,}  with  128GB of RAM.
The running time of the trained decision-making policy~$\mu^*$, including the calculation of the feature vector, is 0.25~seconds on average  and 2 seconds in the worst case.
This is \emph{sufficiently low for our problem setting}, where we typically have a couple of seconds to make an online decision.
The running time of one episode of training is 1 day on average and 2 days in the worst case. 
Note that this running time cannot be significantly lowered (other than simulating multiple episodes in parallel) because the training environment has to simulate the real bookings process, where the anytime algorithm is running for the entire day.
As for the offline VRP solvers, the greedy algorithm $\alpha^\textit{Greedy}$ can assign all trip requests $\vect{R}$ for a day in 15 seconds with tight pickup windows $\vect{w}$ and in 3 minutes with broad pickup windows $\vect{W}$; the simulated annealing algorithm $\alpha^\textit{SimAnn}$ performs around 30 iterations per second;
and VROOM and Google OR-Tools take around 3.3 minutes and 1 minute on average, respectively, to solve an offline VRP~instance.



\pgfplotstableread[col sep=comma,]{data/comparison_2/comparison_vroom_anytime_120.csv}\vroomrrtwo
\pgfplotstableread[col sep=comma,]{data/comparison_2/comparison_vroom_anytime_180.csv}\vroomrrthree
\pgfplotstableread[col sep=comma,]{data/comparison_2/comparison_vroom_anytime_240.csv}\vroomrrfour
\pgfplotstableread[col sep=comma,]{data/comparison_6/comparison_routing_anytime_120.csv}\routingrrtwo
\pgfplotstableread[col sep=comma,]{data/comparison_6/comparison_routing_anytime_180.csv}\routingrrthree
\pgfplotstableread[col sep=comma,]{data/comparison_6/comparison_routing_anytime_240.csv}\routingrrfour

\begin{figure}
\begin{tikzpicture}
\begin{axis}[
      boxplot/draw direction=y,
      xtick={1,2,3},
      xticklabels={{2 hours}, {3 hour}, {4 hour}},
      width=\columnwidth,
      height = 4.5cm,
      ymajorgrids,
      major grid style={draw=gray!25},
      bugsResolvedStyle/.style={},
      ylabel={Reduction in Cost},
      yticklabel=\pgfmathprintnumber{\tick}\,$\%$,
      xlabel={Broad Window Size},
      font=\small,
    ]

\addplot+[boxplot={box extend=0.10, draw position=1},ColorVroom, solid, lshift, fill=ColorVroom!20, mark=x] table [col sep=comma, y=diff_vroom_anytime] {\vroomrrtwo};
\addplot+[boxplot={box extend=0.10, draw position=2}, ColorVroom, solid, lshift, fill=ColorVroom!20, mark=x] table [col sep=comma, y=diff_vroom_anytime] {\vroomrrthree};
\addplot+[boxplot={box extend=0.10, draw position=3}, ColorVroom, solid,lshift, fill=ColorVroom!20, mark=x] table [col sep=comma, y=diff_vroom_anytime] {\vroomrrfour};

\addplot+[boxplot={box extend=0.10, draw position=1},ColorGoogle, solid, rshift, fill=ColorGoogle!20, mark=x] table [col sep=comma, y=diff_routing_anytime] {\routingrrtwo};
\addplot+[boxplot={box extend=0.10, draw position=2}, ColorGoogle, solid, rshift, fill=ColorGoogle!20, mark=x] table [col sep=comma, y=diff_routing_anytime] {\routingrrthree};
\addplot+[boxplot={box extend=0.10, draw position=3}, ColorGoogle, solid,rshift, fill=ColorGoogle!20, mark=x] table [col sep=comma, y=diff_routing_anytime] {\routingrrfour};
\end{axis}
\end{tikzpicture}
\caption{Reduction in total cost due to using our approach 
for selecting tight pickup windows (policy $\mu^*$ supported by anytime algorithm $\alpha^\textit{SimAnn+Greedy}$), compared to using na\"ive pickup windows with VROOM (\textcolor{ColorLegendVroom}{$\blacksquare$}) and Google OR-Tools  (\textcolor{ColorLegendGoogle}{$\blacksquare$}) as offline VRP solvers.  
}
\label{fig:ourbest_vs_otherbaselinee_static_updated}
\end{figure}
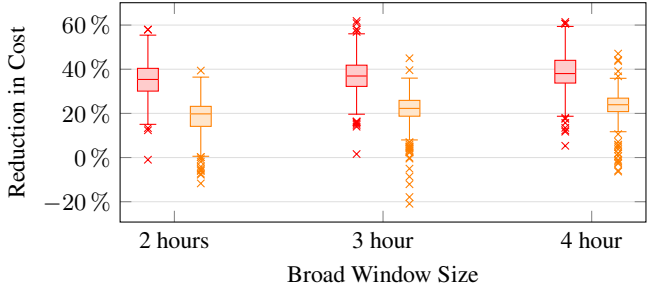

\paragraph{Proposed Approach vs. Na\"ive Pickup Windows}
Next, we demonstrate the effectiveness of our proposed approach by showing that optimizing tight pickup windows can lead to significant reductions in cost.
Since the online bookings problem is novel to the best of our knowledge, existing VRP solvers do not address the online selection of tight pickup windows. 
Therefore, to provide baselines for comparison, we consider existing offline VRP solvers with ``na\"ively selected'' pickup windows, which we define as selecting the middle interval of broad pickup windows as tight windows.

\cref{fig:ourbest_vs_otherbaselinee_static_updated} shows the reduction in the total cost of vehicle routes due to using our proposed approach, compared to using VROOM and Google OR-Tools with na\"ively selected pickup windows.
For each comparison, we evaluate the algorithms on 180 days of paratransit data, and plot the distributions.
We observe a significant reduction in costs compared to both baseline solvers.
Further, we find that our approach is robust to variations in the duration of broad pickup windows since it maintains a significant advantage when broad windows are 2- or 4-hours long, even though the decision policy $\mu^*$ was trained only on 3-hour broad windows.



\pgfplotstableread[col sep=comma,]{data/comparison_2/comparison_rt_sa_120_anytime_120.csv}\ourntwo
\pgfplotstableread[col sep=comma,]{data/comparison_2/comparison_rt_sa_180_anytime_180.csv}\ournthree
\pgfplotstableread[col sep=comma,]{data/comparison_2/comparison_rt_sa_240_anytime_240.csv}\ournfour

\pgfplotstableread[col sep=comma,]{data/comparison_2/comparison_sim_anneal_anytime_120.csv}\ournntwo
\pgfplotstableread[col sep=comma,]{data/comparison_2/comparison_sim_anneal_anytime_180.csv}\ournnthree
\pgfplotstableread[col sep=comma,]{data/comparison_2/comparison_sim_anneal_anytime_240.csv}\ournnfour

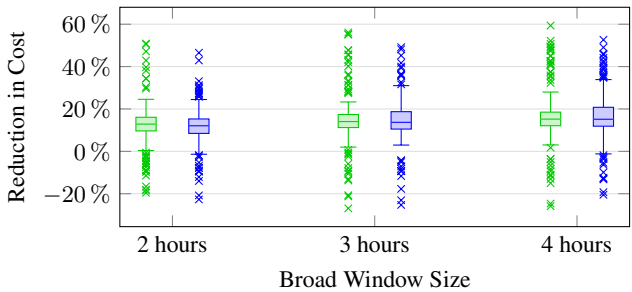
\begin{figure}
\begin{tikzpicture}
\begin{axis}[
      boxplot/draw direction=y,
      xtick={1,2,3},
      xticklabels={{2 hours}, {3 hours}, {4 hours}},
      width=0.975\columnwidth,
      height = 4.5cm,
      ymajorgrids,
      major grid style={draw=gray!25},
      ylabel={Reduction in Cost},
      yticklabel=\pgfmathprintnumber{\tick}\,$\%$,
      xlabel={Broad Window Size},
      font=\small,
    ]

\addplot+[boxplot={box extend=0.10, draw position=1},ColorOur, solid, lshift, fill=ColorOur!20, mark=x] table [col sep=comma, y=diff_rt_sa_anytime] {\ourntwo};
\addplot+[boxplot={box extend=0.10, draw position=2}, ColorOur, solid, lshift, fill=ColorOur!20, mark=x] table [col sep=comma, y=diff_rt_sa_anytime] {\ournthree};
\addplot+[boxplot={box extend=0.10, draw position=3}, ColorOur, solid, lshift, fill=ColorOur!20, mark=x] table [col sep=comma, y=diff_rt_sa_anytime] {\ournfour};

\addplot+[boxplot={box extend=0.10, draw position=1},ColorOurWOS, solid, rshift, fill=ColorOurWOS!20, mark=x] table [col sep=comma, y=diff_sim_anneal_anytime] {\ournntwo};
\addplot+[boxplot={box extend=0.10, draw position=2}, ColorOurWOS, solid,rshift, fill=ColorOurWOS!20, mark=x] table [col sep=comma, y=diff_sim_anneal_anytime] {\ournnthree};
\addplot+[boxplot={box extend=0.10, draw position=3}, ColorOurWOS, solid,rshift, fill=ColorOurWOS!20, mark=x] table [col sep=comma, y=diff_sim_anneal_anytime] {\ournnfour};
\end{axis}
\end{tikzpicture}
\caption{Reduction in total cost due to using our complete approach (policy $\mu^*$ supported by anytime algorithm $\alpha^\textit{SimAnn+Greedy}$), compared to using a policy $\mu^*$ without anytime support (\textcolor{ColorLegendOur}{$\blacksquare$}) and using na\"ive pickup windows with algorithm $\alpha^\textit{SimAnn+Greedy}$ as the offline VRP solver  (\textcolor{ColorLegendOurWOS}{$\blacksquare$}).
}
\label{fig:ourrtat_vs_ourrt_sa_vs_ourg_sa}
\end{figure}

\paragraph{Advantage of Combining Policy with Anytime Algorithm}
While \cref{fig:ourbest_vs_otherbaselinee_static_updated} demonstrates the effectiveness of our proposed approach, it does not prove that every element of our approach is  necessary. 
One may wonder if a simpler approach would work equally well. 
To demonstrate that both the anytime algorithm and the learning-based decision policy are crucial, we compare our complete approach to (1) using the decision policy $\mu^*$ without anytime support and (2) using the anytime algorithms $\alpha^\textit{SimAnn+Greedy}$ as  offline VRP solvers with na\"ive pickup windows (i.e., without decision policy). 

\cref{fig:ourrtat_vs_ourrt_sa_vs_ourg_sa} shows the reduction in the total cost of vehicle routes due to using our complete approach compared to incomplete variants (1) and (2).
We observe that there is a significant reduction in cost compared to both, which demonstrates that both the learning-based decision policy $\mu^*$ and the anytime algorithms $\alpha^\textit{SimAnn+Greedy}$ are crucial.


\newcommand{\citeSubject}[1]{\citeauthor{#1}~\shortcite{#1}}

\section{Related Work}
\label{sec:related}





Some prior works focus on solving the dial-a-ride problem, an online VRP \cite{berbeglia2012hybrid,liu2015branch,parragh2015dial,wilbur2022online}. 


\citeSubject{mo2018mass} focus on advance booking in an offline VRP. We also consider advance bookings (i.e., day before the travel). \citeSubject{de2021integrated} consider enhancing the solution quality of offline VRP by using online algorithms that can optimize the solution obtained from offline VRP algorithms. Prior works such as \cite{lowalekar2019zac,shen2019roo,alonso2017demand,ota2016stars,simonetto2019real,james2019online,joe2020deep} consider real-time demand. Among them, \citeSubject{simonetto2019real} and \citeSubject{alonso2017demand} consider real-time positioning of  vehicles. \citeSubject{gupta2010improving} and \citeSubject{wen2018transit} consider both real-time vehicle scheduling and advance booking. \citeSubject{simonetto2019real} consider a system where the agency uses idle vehicles by relaxing the time-related constraints, rather than rejecting user requests.

\citeSubject{nguyen2019hierarchical} consider a hierarchical approach by prioritizing requests. In our paratransit service setting, we treat all service requests with the same priority. \citeSubject{simonetto2019real} assign one request to one vehicle from a given batch of requests for faster real-time assignment. \citeSubject{goodson2017rollout} consider a lookahead strategy by using rollout algorithms. \citeSubject{joe2020deep} consider a route-based MDP.

\section{Conclusion}

Optimizing pickup windows during day-ahead trip booking can be crucial for offline VRPs (e.g., paratransit service applications). In this paper, we propose a novel problem formulation to capture offline VRPs with online bookings. We also introduce a novel computational approach that combines a learning-based policy with an anytime algorithm. Based on experiments with real-world paratransit data from CARTA, the public transit agency of Chattanooga, TN, 
we observe a significant reduction in costs due to selecting  
pickup windows using our decision policy instead of na\"ive selection. 
Further, our experiments also show a reduction of 14 - 18\% in costs due to using our policy in tandem with an anytime algorithm instead of using the policy by itself. 

\paragraph*{Acknowledgements}
This material is based upon work sponsored by the National Science Foundation under Grant CNS-1952011 and by the Department of Energy under Award DE-EE0009212.
Any opinions, findings, and conclusions or recommendations expressed in this material are those of the authors and do not necessarily reflect the views of the National Science Foundation or the Department of Energy.

\bibliographystyle{named}
\bibliography{main}

\ifExtendedVersion

\clearpage
\appendix

\section{Notation}
\label{app:symbols}
\begin{table}[h!]
\setcounter{section}{1}

    \renewcommand{\arraystretch}{1.4}
    \caption{List of Symbols}
    \begin{tabular}{|c|l|}
        \hline
        Symbol & Description \\
        \hline\hline
        \multicolumn{2}{|c|}{Offline Vehicle Routing Problem (VRP)} \\
        \hline
        $\LocationVector$ &  set of locations\\
        \hline
        $L^\textit{depot}\in \LocationVector$ &  location of the vehicle depot (i.e., garage) \\
        \hline
        $\RequestsVector$ & ordered set of trip requests ($\RequestsVector = \langle T_1, T_2, \ldots \rangle$)\\
        \hline
        $L^{\textit{pickup}}_i \in \LocationVector$ &  pickup location of trip request $T_i \in \RequestsVector$ \\
        \hline
        $L^{\textit{dropoff}}_i \in \LocationVector$ &  drop-off location of trip request $T_i \in \RequestsVector$ \\
        \hline
        $P_i$ &  passenger occupancy of trip request $T_i \in \RequestsVector$ (i.e., number of passengers to be transported) \\
        \hline
        $\vect{w}$ & ordered set of pickup time windows ($\WindowsVector = \langle w_1, w_2, \ldots \rangle$; $|\RequestsVector| = |\WindowsVector|$) \\
        \hline
        $w^\textit{start}_i$ &  start time of pickup window $w_i \in \vect{w}$ (i.e., earliest pickup time) \\
        \hline
        $w^\textit{end}_i$ &  end time of pickup window $w_i \in \vect{w}$ (i.e., latest pickup time) \\
        \hline
        $D^\textit{dwell}$ &  dwell time for pickup and dropoff \\
        \hline
        $D^\textit{maxroute}$ &  maximum duration of a vehicle route \\
        \hline
        $D^{\textit{travel}}_{l_1,l_2}$ &  time to drive from location $l_1 \in \LocationVector$ to location $l_2 \in \LocationVector$\\
        \hline
        $V$ & vehicle passenger capacity (i.e., maximum number of passengers on a vehicle at a time) \\
        \hline
        $C^\textit{nroutes}$ & cost factor for the number of routes in the objective function \\
        \hline\hline
        \multicolumn{2}{|c|}{Offline VRP Solution}\\
        \hline
        $\RunsVector$ &  set of vehicle routes ($\RunsVector = \langle R_1, R_2, \ldots \rangle$)\\
        \hline
        $R_i^\textit{start}$ &  start time of route $R_i \in \RunsVector$ (i.e., time when vehicle leaves the depot) \\
        \hline
        $R_i^\textit{end}$ &  end time of route $R_i \in \RunsVector$ (i.e., time when vehicle returns to the depot) \\
        \hline
        $\Feasible(\RequestsVector, \WindowsVector)$ & set of feasible solution for the VRP instance $(\RequestsVector, \WindowsVector)$ \\
        \hline
        $\Cost(\RunsVector)$ & total cost of the VRP solution $\RunsVector$ \\
        \hline
        $\VRP^*(\RequestsVector, \WindowsVector)$ & total cost of an optimal solution for the VRP instance $(\RequestsVector, \WindowsVector)$ (i.e., $\VRP^*(\RequestsVector, \WindowsVector) = \min_{\vect{R} \in \Feasible(\RequestsVector, \WindowsVector)} \Cost(\vect{R})$)\\
        \hline\hline
        \multicolumn{2}{|c|}{Online Bookings Problem}\\
        \hline
        $\vect{W}$ & ordered set of broad pickup time windows ($\vect{W} = \langle W_1, W_2, \ldots\rangle$; $|\vect{W}| = |\RequestsVector|$) \\
        \hline
        $W^\textit{start}_i$ &  start time of window $W_i \in \vect{W}$ (i.e., earliest pickup time) \\
        \hline
        $W^\textit{end}_i$ &  end time of window $W_i \in \vect{W}$ (i.e., latest pickup time) \\
        \hline
        $\mathcal{D}$ & probability distribution of $(\RequestsVector, \vect{W})$ \\
        \hline
        $D^\textit{window}$ & maximum duration of a tight pickup window \\
        \hline
        $\mu^*$ &  optimal decision policy for the online bookings problem
        \\
        \hline
        \hline
        \multicolumn{2}{|c|}{Solution Approach}\\
        \hline
        $\alpha$ & anytime algorithm for solving an offline VRP with supporting input $\hat{\vect{R}}{}^{(i)}$ \\
        \hline
        $\vect{R}^{(i)}$ & feasible solution (i.e., set of routes) output by the $i$th execution of the anytime algorithm $\alpha$ \\
        \hline
        $\hat{\vect{R}}{}^{(i)}$ & feasible solution (i.e., set of routes) output by the $i$th execution of the decision policy $\mu$ \\
        \hline
        $c_i$ & immediate cost incurred after the $i$th decision (MDP formulation) \\
        \hline
        $\tilde{c}_i$ & shaped immediate cost incurred after the $i$th decision \\
        \hline
        $Q$ & value function for predicting cost $\tilde{c}_i$ for a state-action pair (i.e., for a decision input and decision) \\
        \hline
    \end{tabular}
    \label{table:symbol_list}
\end{table}

\clearpage
\section{Offline Vehicle Routing Problem with Time Windows}
\label{app:offline_problem}
\newcommand{\Natural}[0]{\ensuremath{\mathbb{N}}}
\newcommand{\Real}[0]{\ensuremath{\mathbb{R}}}
\newcommand{\AssignmentVector}[0]{\ensuremath{\vect{A}}}

In \cref{sec:offline_VRP_short}, due to lack of space, we provided a brief description of the \emph{offline vehicle routing problem with time windows}, which was sufficient for formulating the online bookings problem. Here, we provide a complete and formal description of the offline VRP problem. 
Please note that
the online bookings problem can be defined with respect to a range of offline VRP variants that consider pickup time windows, and our proposed solution approach could incorporate any offline VRP solver for these problems.

\subsubsection{Notation}
Throughout the description of the model, we use $\Real$ to denote the set of real numbers, $\mathbb{N}$ to denote the set of natural numbers, $\vect{L}$ to denote the set of locations, and $D^\textit{travel}_{l_1,l_2}$ to denote the travel time between locations $l_1 \in \vect{L}$ and $l_2 \in \vect{L}$.
We assume that points in time as well as time durations are represented by real numbers. \ad{positive real numbers? second since midnight?}

\subsubsection{Input}
The input of the offline VRP problem is
\begin{itemize}
    \item an ordered set of \emph{trip requests} $\RequestsVector = \langle T_1, T_2, \ldots, T_n\rangle$, where each trip request $T_i$ contains a pickup location $L^{\textit{pickup}}_i \in \vect{L}$, a drop-off location $L^{\textit{dropoff}}_i \in \vect{L}$, and the number of passengers to be transported $P_i$;
    \item a corresponding ordered set of \emph{tight pickup time windows} $\WindowsVector = \langle w_1, w_2, \ldots, w_n\rangle$, where each time window $w_i$ is defined by an earliest $w_i^\textit{start} \in \mathbb{R}$ and latest $w_i^\textit{end} \in \mathbb{R}$ pickup time;
    \item the maximum allowed duration $D^\textit{maxroute} \in \mathbb{R}$ of a vehicle route (e.g., maximum length of a driver's shift); 
    \item the passenger capacity $V \in \mathbb{N}$ of a vehicle (i.e., maximum number of passengers on board at a time);
   \item the dwell time $D^\textit{dwell} \in \Real$ at the pickup and drop-off locations;
   \item the location $L^\textit{depot} \in \LocationVector$ of the vehicle depot (i.e., garage);
   \item the cost factor $C^\textit{nroutes}$ for the number of routes in the objective function.
\end{itemize}
For ease of presentation, we  represent a VRP instance as $(\RequestsVector, \WindowsVector)$, assuming that the constants are provided implicitly.

\subsubsection{Solution Representation}

A solution to the offline VRP problem is
a set of \emph{vehicle routes} $\vect{R} = \{ R_1, R_2, \ldots, R_m\}$.
Informally, each vehicle route is a list of locations (pickup or drop-off) with  associated arrival times (i.e., when the vehicle arrives at the location to pick up or drop off passengers).
Formally, a vehicle route $R_i$ is an ordered set of tuples $\langle l, t \rangle \in \vect{L} \times \mathbb{R}$, where $l$ is either a pickup  $L_j^\textit{pickup}$ or drop-off location $L_j^\textit{dropoff}$, and $t$ is the time when the vehicle on route $R_i$ arrives at location $l$.
Note that in \cref{sec:offline_VRP_short}, we represented a vehicle route as an ordered set of pickup and drop-off locations for ease of exposition. While we could use that representation here, it will actually be easier to use this representation to provide a complete formal definition of the offline VRP problem.


\subsubsection{Constraints}
A set of vehicle routes $\vect{R}$ is a feasible solution to  the offline VRP problem $(\RequestsVector, \WindowsVector)$ if it satisfies the following set of constraints.

\Aron{not strictly necessary, but it may be more clear to the reader if we include this (also, we mentioned this in the main text)}
First, each  trip $T_i \in \RequestsVector$ is picked up by at most one vehicle route $R_j \in \vect{R}$:
\begin{align*}
& \forall T_i \in \vect{T}, R_j \in \vect{R}, t_j \in \mathbb{R}, R_k \in \vect{R}, t_k \in \mathbb{R}: \nonumber\\
& \langle L_i^\textit{pickup}, t_j \rangle \in R_j \wedge \langle L_i^\textit{pickup}, t_k \rangle \in R_k \Rightarrow j = k \wedge t_j = t_k 
\end{align*}
In other words, if trip $T_i$ is picked up by route $R_j$ at $t_j$, then no other route $R_k$ can pick up this trip (and neither can this route $R_j$ at any other time $t_k$).
Similarly, each trip $T_i \in \RequestsVector$ is dropped off by at most one route $R_j \in \vect{R}$:
\begin{align*}
& \forall T_i \in \vect{T}, R_j \in \vect{R}, t_j \in \mathbb{R}, R_k \in \vect{R}, t_k \in \mathbb{R}: \nonumber\\
& \langle L_i^\textit{dropoff}, t_j \rangle \in R_j \wedge \langle L_i^\textit{dropoff}, t_k \rangle \in R_k \Rightarrow j = k \wedge t_j = t_k 
\end{align*}

Second, each trip  $T_i \in \RequestsVector$ is served by at least one route $R_j \in \vect{R}$ such that the passengers are picked up by the vehicle within the time window $w_i$ and dropped off on time, by $w_i^\textit{end} + D^\textit{travel}_{L^\textit{pickup}_i, L^\textit{dropoff}_i}$ at latest:
\begin{align*}
        \forall T_i \in \RequestsVector: \\
        && \exists R_j \in & ~ \RunsVector,  t^\textit{pickup} \in \Real, t^\textit{dropoff} \in \Real \wedge \Big( & \nonumber \\
        &&& \langle L^\textit{pickup}_i, t^\textit{pickup} \rangle , \langle L^\textit{dropoff}_i, t^\textit{dropoff} \rangle \in R_j  \nonumber \\
        && \wedge ~ & w^\textit{start}_i \leq t^\textit{pickup} \leq w^\textit{end}_i  \nonumber \\
        && \wedge ~ & t^\textit{pickup} < t^\textit{dropoff}  \nonumber \\
        && \wedge ~ & t^\textit{dropoff} \leq  w^\textit{end}_i + D^\textit{travel}_{L^\textit{pickup}_i, L^\textit{dropoff}_i} \nonumber \Big) 
    \end{align*}
In other words, for each trip request $T_i \in \vect{T}$, there exists a vehicle run $R_j \in \vect{R}$ that picks up the passengers at some time $t^\textit{pickup} \in \mathbb{R}$ and drops them off at some time $t^\textit{dropoff} \in \mathbb{R}$ (second and third lines),
the pickup time $t^\textit{pickup}$ is within the pickup window $w_i$ (fourth line),
the pick time  $t^\textit{pickup}$ is earlier thn the drop-off time $t^\textit{dropoff}$ (fifth line),
and the drop-off time $t^\textit{dropoff}$ is no later than $w_i^\textit{end} + D^\textit{travel}_{L^\textit{pickup}_i, L_i^\textit{dropoff}}$ (sixth line).
Note that the last clause ensures that passengers arrive at their dropoff location $L_i^\textit{dropoff}$ no later than if they were picked up at the latest possible time $w_i^\text{end}$ and drove to the dropoff location without any detours.
Our formulation and algorithms could very easily be extended to consider detour times (defined with respect to pickup times).
In our experiments, we consider the above formulation since it captures the requirements of the transit agency.


Third, every vehicle route $R_j \in \vect{R}$ satisfies travel-time and dwell-time constraints:
\begin{align*}
        \forall R_j \in \RunsVector,l_1 \in \LocationVector,  l_2 \in \LocationVector, &~ t_1 \in \Real, t_2 \in \Real: \nonumber \\
        & \langle l_1, t_1 \rangle, \langle l_2, t_2 \rangle \in {R_j} \wedge t_1 < t_2 \nonumber \\
        & \Rightarrow t_1 + D^\textit{dwell} + D^\textit{travel}_{l_1, l_2} \leq t_2
\end{align*}
In other words, if vehicle route $R_j \in \vect{R}$ arrives at location $l_1 \in \vect{L}$ at time $t_1 \in \mathbb{R}$ and later arrives at location $l_2 \in \vect{L}$ at time $t_2 \in \mathbb{R}$, then the time difference between $t_1$ and $t_2$ must be at least $D^\textit{dwell}$ (time required to pick up or drop off passengers at location $l_1$) plus $D^\textit{travel}_{l_1, l_2}$ (time required to drive from location $l_1$ to $l_2$).
Note that if the constraint is satisfied for consecutive locations, then it is also satisfied for non-consecutive ones due to the triangle inequality, so the above constraint could be reformulated to consider only consecutive locations. We use the above formulation for the sake of simplicity.

Fourth, the duration of each vehicle route $R_j \in \vect{R}$ is at most $D^\textit{maxroute}$, where the duration of a route is the time between leaving the garage (i.e., vehicle depot) and returning to it.
We introduce this constraint based on input from the transit agency, which stores all vehicles at a garage overnight, and has strict constraints on the duration of the routes due to the drivers' labour contracts.
To express this, we define the start time $R_j^\textit{start}$ of route $R_j$ as the time when the vehicle needs to leave the garage to serve its first trip:
\begin{align*}
   R_j^\textit{start} = \min_{\langle l, t \rangle \in R_j} t - D^\textit{travel}_{L^\textit{depot}, l} 
\end{align*}
Similarly, we define the end time $R_j^\textit{end}$ of route $R_j$ as the time when the vehicle can arrive at the garage after serving its last~trip:
\begin{align*}
   R_j^\textit{end} = \max_{\langle l, t \rangle \in R_j} t + D^\textit{dwell} + D^\textit{travel}_{l,L^\textit{depot}} 
\end{align*}
Then, we can formulate the constraint on the duration of vehicle routes as follows:
    \begin{align*}
        \forall R_j \in \RunsVector: ~~~ R^\textit{end}_j \leq R^\textit{start}_j + D^\textit{maxroute}
    \end{align*}

Finally, the number of passengers on board a vehicle does not exceed the passenger capacity $V$ of a vehicle at any time.
When vehicle route $R_j \in \RunsVector$ picks up passengers at a pickup location $L^\textit{pickup}_i$, the occupancy of the vehicle serving route $R_j$ is incremented by the number of passengers  $P_i$.
Similarly, when  route $R_j \in \RunsVector$ drops off  passengers at a drop-off location $L^\textit{dropoff}_i$, the occupancy of the vehicle serving  route $R_j$ is decreased by the number of passengers $P_i$. 
For each route $R_j \in \RunsVector$, the occupancy of the vehicle serving the route at any time $t \in \mathbb{R}$ is less than or equal to the passenger capacity~$V$:
    \begin{align*}
        \forall R_j \in \RunsVector, t \in \Real: &~  \nonumber \\
        & \sum_{T_i \in \RequestsVector, t' \in \Real: ~\langle L^\textit{pickup}_i, t' \rangle \in {R_j} \wedge t' < t} P_i \nonumber \\
        & - \sum_{T_i \in \RequestsVector, t' \in \Real: ~\langle L^\textit{dropoff}_i, t' \rangle \in {R_j} \wedge t' < t} P_i
        \leq V
    \end{align*}
Note that the first summation adds up all the passengers who have been picked up before (at some time $t' < t$), while the second summation adds up all the passengers who have been dropped up before (at some time $t' < t$).
Hence, their difference is the number of passengers on board the vehicle at time~$t$.

\subsubsection{Objective} 
We define the objective of the offline VRP as minimizing the total cost of the vehicle routes $\vect{R}$, which depends on the duration and number of vehicle routes.
Formally, we define the total cost $\Cost(\vect{R})$ of a set of vehicle routes~$\vect{R}$ as follows:
\begin{align}
   \Cost(\RunsVector) =  \sum_{R_j \,\in\, \RunsVector} (R^{\textit{end}}_j - R^{\textit{start}}_j) + C^\textit{nroutes} \cdot |\RunsVector|
\end{align}
where $C^\textit{nroutes}$ is a cost factor that captures the constant ``overhead'' costs associated with each vehicle route.
In practice, an 8-hour vehicle route does not cost eight times as much as a 1-hour vehicle route since there are constant costs associated with the route (e.g., preparing a vehicle for the drive, bringing in a driver, or markup for outsourcing).
In fact, very short routes (e.g., 20 minutes) may be prohibitively uneconomical in practice.
The second term of the offline VRP objective enables capturing this.

\clearpage
\section{Features}
\label{app:features}
\newcommand{\Busyness}[0]{\ensuremath{\mathcal{BN}}}
\newcommand{\ExpReq}[0]{\ensuremath{\mathcal{ER}}}

In this section, we provide a detailed description of the features that the value function $Q$ takes as input, which we briefly introduced in \cref{sec:decision_policy}.
Please recall that the values of these features are defined for a particular state (i.e., decision input) $(\RequestsSubVector{i}, \WindowsSubVector{i-1}, W_i, \vect{R}{}^{(i-1)})$ and particular action (i.e., decision) $(w_i, \hat{\vect{R}}{}^{(i)})$.

First, we describe features based on the probability distribution $\mathcal{D}$, which capture our expectations of future requests. 

\paragraph{Busyness ($\Busyness$)} 
These features consider how busy certain locations and certain times of the day are, in terms of the number of expected trip requests around those locations and times of day.
Specifically, they consider the expected number of daily trip requests whose pickup locations $L_j^\textit{pickup}$ are in the same geographical area as the pickup location $L_i^\textit{pickup}$ of request $T_i$ (e.g., in the same ZIP-code area in the U.S.) and/or whose drop-off locations $L_j^\textit{dropoff}$ are in the same geographical area as the drop-off location $L_i^\textit{dropoff}$ and/or whose broad pickup windows start around the same time  $W_j^\textit{start}$ as the narrow window $w_i^\textit{start}$ (e.g., within the same 1-hour interval).
In other words, these features consider the expected number of trip requests received for a day that satisfy some of the following criteria:
\begin{itemize}
    \item same geographical area as the pickup location $L^\textit{pickup}_i$,
    \item same geographical area as the drop-off location $L^\textit{dropoff}_i$,
    \item similar time as the tight pickup windows $w_i$.
\end{itemize}

Based on the above three criteria, we define four variants of the \emph{busyness feature}:
\begin{itemize}
    \item $\Busyness(\mathcal{D}, L^{\textit{pickup}}_i, L^{\textit{dropoff}}_i, w_i)$: expected number of daily trip requests that travel from the geographical area of $L^\textit{pickup}_i$ to the geographical area of $L^\textit{dropoff}_i$ and whose broad pickup windows start around the same time as $w_i$.

    \item 
  $\Busyness(\mathcal{D}, L^{\textit{pickup}}_i, w_i)$: expected number of daily trip requests that travel from the geographical area of $L^\textit{pickup}_i$ and whose broad pickup windows start around the same time as $w_i$. 
    \item
 $\Busyness(\mathcal{D}, L^{\textit{dropoff}}_i, w_i)$: expected number of daily trip requests that travel  to the geographical area of $L^\textit{dropoff}_i$ and whose broad pickup windows start around the same time as $w_i$. 
    \item
 $\Busyness(\mathcal{D}, w_i)$: expected number of daily trip requests  whose broad pickup windows start around the same time as $w_i$. 
\end{itemize}

All of the above features can be estimated based on historical data (i.e., using an empirical distribution for $\mathcal{D}$) for any given state-action pair.
Note that the distribution of requests~$\mathcal{D}$ may vary significantly between days (e.g., weekends are typically less busy than weekdays); hence, we can use a different $\mathcal{D}$ depending on the day of week.


\paragraph{Expected Requests ($\ExpReq$)}
While the above features consider the expected number of requests received in a whole day, it is also helpful to consider how many \emph{more} requests we expect to receive for the day.
To capture this, we introduce the \emph{expected requests} $\ExpReq(\mathcal{D}, i)$ feature, which is the expected value of $|\RequestsVector| - i$ when making the $i$th decision.
Note that in practice, this feature can also depend on the day of week since the distribution $\mathcal{D}$ varies among the days.
Further, since requests do not arrive at the same rate throughout the day (i.e., during some hours of the day, the agency receives many more calls to book trips than during other hours), we also consider the time of day when the booking call is received to estimate $\ExpReq$ (based again on historical data, in this case the rate at which booking calls are received throughout the day).





\subsubsection{State Features}
Finally, besides considering future requests, we must also consider trip requests that we have already been received.
To this end, we introduce three features that capture how well a decision  
$(w_i, \hat{\vect{R}}{}^{(i)})$ fits the previously selected tight windows $\WindowsSubVector{i-1}$ and vehicle routes~$\RunsVector{}^{(i-1)}$: 
\begin{itemize}
    \item \emph{Time increase} $\mathcal{TI}(\hat{\vect{R}}{}^{(i)}, \RunsVector{}^{(i-1)})$: increase in the duration of the routes, compared between $\hat{\vect{R}}{}^{(i)}$ and $\RunsVector{}^{(i-1)}$:
    \begin{align*}
        \mathcal{DI}(\RunsVector{}^{(i-1)}, \hat{\vect{R}}{}^{(i)}) = & \sum_{R_j \in \hat{\vect{R}}{}^{(i)}} \left( R_j^\textit{end} - R_j^\textit{start}\right) \\ & - \sum_{R_j \in \RunsVector{}^{(i-1)}} \left( R_j^\textit{end} - R_j^\textit{start}\right)
    \end{align*}
    \item \emph{Distance increase} $\mathcal{DI}(\hat{\vect{R}}{}^{(i)}, \RunsVector{}^{(i-1)})$: increase in the driving distances of the routes, compared between $\hat{\vect{R}}{}^{(i)}$ and $\RunsVector{}^{(i-1)}$ (i.e., same as $\mathcal{TI}$, but considering distance driven instead of time spent).
    \item \emph{Tightness of schedule} $\mathcal{TS}(\hat{\vect{R}}{}^{(i)})$: tightness of the route schedule captures how well trip  $T_i$ fits into route $R_j$, where $R_j$ is the route serving trip $T_i$ in  solution $\hat{\vect{R}}{}^{(i)}$.
    The rationale behind this feature is to express how ``fragmented'' a route schedule is, i.e., how much waiting time there is between consecutive trips, which is not long enough to allow serving another trip, but long enough to significantly increase route duration.
    To capture this, we consider the amount of waiting time $x$ between trip $T_i$ and the preceding trip on route $R_j$ (i.e., waiting time before trip $T_i$  on route $R_j$), and the amount of waiting time $y$ between trip $T_i$ and the following trip (i.e., waiting time after serving trip $T_i$), where the waiting time between two consecutive trips is defined as the amount of time between dropping off all passengers from the previous trip and needing to leave for the next trip.
    Note that when consecutive trips are interleaved, waiting time is defined to be zero.
    Then, we can formulate the \emph{tightness of schedule} feature as $\mathcal{TS}(\hat{\vect{R}}{}^{(i)}) = \frac{|x - y|}{|x + y|}$.
    By maximizing this feature, we ensure that time gaps in the route schedule are either minimized (to avoid waiting) or maximized (so that another trip can be served in the gap). 
\end{itemize}

\clearpage
\section{Simulated-Annealing and Greedy Algorithms}
\label{app:anytime_algorithms}

We first provide a high-level overview of the simulated-annealing and greedy algorithms in \cref{sec:alg_overview} and then describe them in detail in \cref{sec:simann} and \cref{sec:greedy}, respectively.

\subsection{Overview of Algorithms}
\label{sec:alg_overview}

\subsubsection{Simulated Annealing}

First, we introduce a simulated annealing algorithm $\alpha^\textit{SimAnn}$, which improves upon a given feasible solution $\hat{\vect{R}}{}^{(i)}$ using an iterative random search.
We provide a detailed description of this algorithm in \cref{app:anytime_algorithms}.
In each iteration, the algorithm 
tries to improve upon the current solution $\vect{R}$ by generating a random neighbor $\vect{R}'$ of the current solution $\vect{R}$ using two operations. 
The first one, called \emph{Swap}, randomly chooses two vehicle routes $R_x, R_y \in \vect{R}$ that overlap in time, and tries to swap a pair of randomly chosen trip request $T_i, T_j$ (where $L_i^\textit{pickup} \in R_x$, $L_j^\textit{pickup} \in R_y$) between the two routes.  
The second one, called \emph{SplitAndMerge}, 
also chooses two overlapping routes $R_x, R_y \in \vect{R}$ at random, but then splits each route into two halves (earlier trips in the first half, later ones in the second) and tries to merge the first half of $R_x$ with the second half of $R_y$ and vice versa.
In each iteration, the algorithm repeatedly applies these operations to the current solution $\vect{R}$  
to obtain a feasible random neighbor $\vect{R}'$. 
Whether this random neighbor replaces the current solution ($\vect{R} \leftarrow \vect{R}'$) or if it is discarded is decided at random, with a probability that depends on $\Cost(\vect{R}) - \Cost(\vect{R}')$.
When terminated, the algorithm returns the best feasible solution that it has encountered during the search as the VRP solution~$\vect{R}{}^{(i)}$.

\subsubsection{Greedy Algorithm}
To enhance the practical performance of our approach, we also introduce a greedy algorithm~$\alpha^\textit{Greedy}$ for solving the paratransit VRP formulation.
While this is not an anytime algorithm, its running time is low enough so that we can successfully execute it between most consecutive requests.
The solution output by the greedy algorithm $\alpha^\textit{Greedy}$ can then be fed into the simulated annealing $\alpha^\textit{SimAnn}$ (if it is better than the current solution).
This algorithm also follows an iterative approach: starting with an empty solution $\vect{R} = \emptyset$, it adds a new routes to the solution one-by-one. For each route, it starts with an empty set  of requests $R= \emptyset$, and tries to assign unserved requests to this route one-by-one, always choosing one that minimizes a heuristic cost function, until there are no feasible assignments or the minimum cost exceeds a threshold.
We provide a detailed description of this algorithm in \cref{app:anytime_algorithms}.

\subsection{Simulated Annealing}
\label{sec:simann}

\begin{algorithm}[!h]
 \caption{$\textbf{Simulated Annealing}(\RunsVector, t_{\textit{run}}, \newline p_{\textit{start}}, p_{\textit{end}}, p_{\textit{alter}},  L^{depot}, D^{\textit{dwell}}, D^{\textit{maxroute}}, V)$}
 \label{algo:simulatedannealing}

$\textnormal{\textit{Solutions}} \leftarrow \{ \RunsVector \}$

$H^{\textit{start}} \leftarrow \frac{-1}{\ln{p_{\textit{start}}}}$

$H^{\textit{end}} \leftarrow \frac{-1}{\ln{p_{\textit{end}}}}$

$H^{\textit{rate}} \leftarrow \left(\frac{H^{\textit{end}}}{H^{\textit{start}}}\right)^{\frac{1}{t_{\textit{run}} -1}} $

$H^{t} \leftarrow H^{\textit{start}}$

$\delta_{avg} \leftarrow 0$


$t^{\textit{start}} \leftarrow \textbf{GetCurrentTime}()$

$t^{\textit{current}} \leftarrow t^{\textit{start}}$

 \While {$t^{\textit{current}} - t^{\textit{start}} \leq t_{\textit{run}}$}
 {  
    
    $\RunsVector' \leftarrow \textbf{RandomNeighbor}(\RunsVector, p_{\textit{alter}})$

    $\delta_e \leftarrow \Cost(\RunsVector') - \Cost(\RunsVector)$
    
    \If{$t^{\textit{current}} - t^{\textit{start}}$ = 1}
    {
        $\delta_{avg} \leftarrow \delta_e$
    }
    
    $\textit{AcceptProbability} \leftarrow \exp{\left(\frac{-\delta_e}{\delta_{avg} \cdot H^t}\right)}$

    \If{$\textnormal{\Cost}(\RunsVector') < \textnormal{\Cost}(\RunsVector)  ~\textnormal{\textbf{or}}~ \textit{AcceptProbability} > \textnormal{\textbf{UniformRandom}}([0,1])$}
    {
    
     $\RunsVector \leftarrow  \RunsVector'$
     
    $\delta_{avg} \leftarrow \delta_{avg} + \frac{\delta_e - \delta_{avg}}{|Solutions|} $
          
     $\textnormal{\textit{Solutions}} \leftarrow \textnormal{\textit{Solutions}} \cup \{\RunsVector\}$

    }
     
     $t^{\textit{current}} \leftarrow \textbf{GetCurrentTime}()$

     $H^{t} \leftarrow H^{\textit{start}} \cdot {H^{\textit{rate}}}^{\{t^{\textit{current}} - t^{\textit{start}}\}}$   
 }
 
 $\RunsVector^* \gets \argmin_{\RunsVector'\in \textnormal{\textit{Solutions}}} \Cost(\RunsVector')$
 
 \KwResult{$\RunsVector^*$} %
\end{algorithm}

The simulated-annealing algorithm follows an iterative process. In each iteration, the algorithm obtains a random neighboring solution $\RunsVector'$ of the current feasible solution $\RunsVector$ using \textbf{RandomNeighbor}. If the total cost ($\Cost$) of $\RunsVector'$ is lower than the total cost of $\RunsVector$, then the algorithm always accepts $\RunsVector'$ as the new current solution. Otherwise, the algorithm computes the probability $\textit{AcceptProbability}$ of accepting $\RunsVector'$ based on the cost difference between $\RunsVector'$ and $\RunsVector$ a decreasing temperature value $H^t$, and then accepts $\RunsVector'$ at random.
Initially, the temperature $H^t$ and probability values are very high (i.e., we initialize $H^0 = H^{\textit{start}}$ with a very high value), which helps the local search to avoid ``getting stuck'' in local optima; over time, the temperature and probability values decrease (i.e., $H^t$ decreases as $t$ increases), enabling the search to converge to an optimum.
The algorithm terminates after its total running time  $(t^{\textit{current}} - t^{\textit{start}})$ exceeds the configured maximum running time $t^{\textit{run}}$, and returns the best solution found up to that point. 

\begin{algorithm}[ht]
 \caption{$\textbf{RandomNeighbor}(\RunsVector, p_{\textit{alter}},  L^{depot}, D^{\textit{dwell}}, \newline D^{\textit{maxroute}}, V)$}
 \label{algo:randomneighbor}

     $\textit{NumberOfAlterations} \leftarrow \textbf{max}\{1, |\RunsVector| \cdot p_{\textit{alter}}\}$
     
    \For {$1, \ldots, \textnormal{\textit{NumberOfAlterations}}$}
    {
     $operation \leftarrow \textbf{UniformRandom}([Swap,SplitAndMerge])$
     
     $\RunsVector \leftarrow operation(\RunsVector)$

    }
    
 \KwResult{$\RunsVector$}
\end{algorithm}

\cref{algo:randomneighbor} follows an iterative process: in each iteration, the algorithm randomly chooses one operation from \emph{Swap} and \emph{SplitAndMerge}, and modifies the input solution $\RunsVector$. The detailed descriptions of the two operations is in the next paragraphs. The running time of this algorithm is
$\mathcal{O}\left(|\RequestsVector|^{4}\right)$.

\begin{algorithm}[ht]
\caption{$\textbf{Swap}(\RunsVector,  L^{depot}, D^{\textit{dwell}}, D^{\textit{maxroute}}, V)$}
\label{algo:swap}

$R_1, R_2 \leftarrow \textbf{UniformRandom}(\RunsVector)$

$T_1 \leftarrow \textbf{UniformRandom}(R_1)$

$T_2 \leftarrow \textbf{UniformRandom}(R_2)$

$R_1^{'}, \textit{SolCost}_1 \leftarrow$ \textbf{Feasible}($R_1, T_2, L^{\textit{depot}}, D^{\textit{dwell}}, D^{\textit{maxroute}}, V$)

 $R_2^{'}, \textit{SolCost}_2 \leftarrow$ \textbf{Feasible}($R_2, T_1, L^{\textit{depot}}, D^{\textit{dwell}}, D^{\textit{maxroute}}, V$)

\If {$\textit{SolCost}_1 \neq \infty \land \textit{SolCost}_2 \neq \infty$}
{
    $\RunsVector \leftarrow \RunsVector \setminus \{R_1, R_2\} \cup \{R_1^{'}, R_2^{'}\} $
}
\KwResult{$\RunsVector$}
\end{algorithm}

\cref{algo:swap} randomly chooses two vehicle routes $R_1, R_2 \in \RunsVector$ that overlap in time, and tries to swap a pair of randomly chosen trip request $T_1, T_2$ (where $L_1^\textit{pickup} \in R_1$, $L_2^\textit{pickup} \in R_2$) between the two routes. If the swap is feasible (i.e., satisfies time and occupancy constraints), then update $\RunsVector$ with the newly modified routes ($R_1', R_2'$) and remove the initial routes ($R_1, R_2$). Finally, return the updated routes $\RunsVector$. 
The time complexity of this algorithm is
$\mathcal{O}\left(|\RequestsVector|^{3}\right)$.

\begin{algorithm}[ht]
\caption{$\textbf{SplitAndMerge}(\RunsVector,  L^{depot}, D^{\textit{dwell}}, \newline D^{\textit{maxroute}}, V)$}
\label{algo:splitandmerge}

$R_1, R_2 \leftarrow \textbf{UniformRandom}(\RunsVector)$

$R_1^1, R_1^2 \leftarrow \textbf{SplitRuns}(R_1)$

$R_2^1, R_2^2 \leftarrow \textbf{SplitRuns}(R_2)$

success$_1$, $R_1^{'} \leftarrow$ \textbf{MergeRuns}($R_1^1, R_2^2, L^{\textit{depot}}, D^{\textit{dwell}}, D^{\textit{maxroute}}, V$)

success$_2$, $R_2^{'} \leftarrow$ \textbf{MergeRuns}($R_2^1, R_1^2, L^{\textit{depot}}, D^{\textit{dwell}}, D^{\textit{maxroute}}, V$)

\If {success$_1 \land$ success$_2$}
{
    $\RunsVector \leftarrow \RunsVector \setminus \{R_1, R_2\} \cup \{R_1^{'}, R_2^{'}\} $
}
\KwResult{$\RunsVector$}
\end{algorithm}

\cref{algo:splitandmerge} chooses two overlapping routes $R_1, R_2 \in \RunsVector$ at random, and then splits each route into two halves (earlier trips in the first half $R_i^1$, later ones in the second half $R_i^2$). Then, the algorithm obtains the merged route $R_1'$ by merging $R_1^1$ and $R_2^2$, and similarly obtains the merged route $R_2'$ by merging $R_2^1$ and $R_1^2$. If both merged routes are feasible, then update $\RunsVector$ with newly modified routes ($R_1', R_2'$) and remove the initial routes ($R_1, R_2$). Finally, return the updated routes $\RunsVector$. The algorithm \textbf{MergeRuns} has the time complexity of $\mathcal{O}\left(|\RequestsVector|^{4}\right)$.
Accordingly, the time complexity of this algorithm is
$\mathcal{O}\left(|\RequestsVector|^{4}\right)$.

\subsection{Greedy Algorithm}
\label{sec:greedy}

\subsubsection{Inputs}

In our VRP formulation (\cref{app:offline_problem}), we defined time windows only for pickups since the latest drop-off time is implicitly defined by the latest pickup time and the travel time for each trip request.
Here, we present a greedy algorithm that can solve a more general VRP formulation, in which time windows are defined for both pickups and drop-offs.
Specifically, we present a greedy algorithm that takes as input---in addition to the inputs defined in \cref{app:offline_problem}---an ordered set of drop-off time windows $\vect{dw}$, where $dw_i^\textit{start}$ is the earliest time that passengers from trip $T_i$ can be dropped off and $dw_i^\textit{end}$ is the latest time that they can be dropped off.
Then, we can solve instances of our problem formulation from \cref{app:offline_problem} by letting $dw_i^\textit{start} = w_i^{\textit{start}} + D^{\textit{travel}}_{L^{\textit{pickup}}_i,L^{\textit{dropoff}}_i}$ and $dw^\textit{end}_i  = w_i^{\textit{end}} + D^{\textit{travel}}_{L^{\textit{pickup}}_i,L^{\textit{dropoff}}_i}$ for each trip request $T_i$.


\subsubsection{Outputs}
The output of the algorithm is  a set of routes $\RunsVector$, as described in \cref{app:offline_problem}. 
To facilitate the description of our algorithms, we extend this solution representation with additional fields.
First, we let the term \emph{node} refer to a pair $\langle l, w \rangle$, where $l \in \LocationVector$ is a location and $w \in \WindowsVector \cup \vect{dw}$ is a time window.

We let each route $R_k \in \RunsVector$ consist of the following:
\begin{itemize}
    \item  $Y_{R_k}$: list of nodes served by route $R_k$. We let $Y_{R_k}^{\textit{first}}$ denote the first node in $Y_{R_k}$ and let $Y_{R_k}^{\textit{last}}$ denote the last node in~$Y_{R_k}$.
    \item $D_{R_k}$: list of points in time ($|D_{R_k}| = |Y_{R_k}|$), where each element represents the time when  route  $R_k$ reaches the location of the corresponding node of $Y_{R_k}$.
    We let $D_{R_k}^{\textit{first}}$ denote the first time point in $D_{R_k}$ and let $D_{R_k}^{\textit{last}}$ denote the last time point in $D_{R_k}$.
    \item $O_{R_k}$: list of integers ($|O_{R_k}| = |Y_{R_k}|$),
    where each element represents the occupancy of the vehicle (i.e., number of passengers on board) when route $R_k$ reaches the location of the corresponding node of $Y_{R_k}$. 
\end{itemize}




\begin{algorithm}[!h]
 \caption{$\textbf{GetPlacements}(R_i, l_n, w_n, \textit{\textit{index}}, D^{\textit{dwell}})$}
 \label{algo:getplacements}
 
$Placements \leftarrow \emptyset$

\For {$\langle l_k, w_k \rangle \in Y_{R_i}$}
{
   
    \If{$k \geq \textit{index}$}{
    
    \If {$k = 0 \land$ \textbf{Reachable}$(l_n, w_n, l_k, w_k, D^{\textit{dwell}})$}
    {
    
        $Placements \leftarrow Placements \cup \{0\}$
        
    }
    
    \uIf {$k < \left|Y_{R_i}\right| - 1$}
    {
        $l_{k+1}, w_{k+1} \leftarrow Y_{R_i}[k + 1]$

        \If {\textbf{Reachable}$(l_k, w_k, l_n, w_n, D^{\textit{dwell}}) \land$ \textbf{Reachable}$(l_n, w_n, l_{k+1}, w_{k+1}, D^{\textit{dwell}})$}
        {
        
        $Placements \leftarrow Placements \cup \{k + 1\}$
         
        }
        
    }
    \ElseIf {$k = \left|Y_{R_i}\right| - 1$}
    {
    
        \If {\textbf{Reachable}$(l_n, w_n, l_k, w_k, D^{\textit{dwell}})$}
        {
        
        $Placements \leftarrow Placements \cup \{k + 1\}$
          
        }
        
    }
    
    }

}
\KwResult{Placements}
\end{algorithm}

\cref{algo:getplacements} identifies the time-feasible placements  of a new node $\langle l_n, w_n \rangle$ into the existing list of nodes $Y_{R_i}$ of  route $R_i$ 
(i.e., positions at which at the new node $\langle l_n, w_n \rangle$ could be inserted into the list $Y_{R_i}$ without violating any time constraints). 
The implementation of the algorithm considers three cases:
\begin{itemize}
    \item First case: the algorithm checks whether the new node $\langle l_n, w_n \rangle$ can be placed before the first node of route $R_i$. In other words, it checks whether the vehicle route $R_i$ can  reach location $l_k \in \LocationVector$ from  location $l_n \in \LocationVector$ without violating any travel-time and time-window constraints.
    
    \item Second case: the algorithm checks for each pair of consecutive nodes $Y_{R_i}[k]$ and $Y_{R_i}[k+1]$ whether it can place the new node $\langle l_n, w_n \rangle$ between  nodes $Y_{R_i}[k]$ and $Y_{R_i}[k+1]$. In other words, it checks  whether vehicle route $R_i$ can  reach  location $l_n \in \LocationVector$ from  location $l_k \in \LocationVector$ and  reach  location $l_{k+1} \in \LocationVector$ from  location $l_n \in \LocationVector$ without violating any travel-time and time-window constraints. 

    \item Third case: 
     The algorithm checks whether the new node $\langle l_n, w_n \rangle$ can be placed as the last node into the route $R_i$.  In other words, it checks  whether vehicle route $R_i$ can reach  location $l_n \in \LocationVector$ from the location $l_k \in \LocationVector$ without violating any travel-time and time-window constraints.
\end{itemize}
The running time of this algorithm is $\mathcal{O}(\left|\RequestsVector\right|)$.

\begin{algorithm}[!h]
 \caption{$\textbf{Feasible}(R_i, T_j, \textit{ratio}, L^{\textit{depot}}, D^{\textit{dwell}},\newline D^{\textit{maxroute}}, V)$}
 \label{algo:feasible}

    $success \leftarrow False$
        
    $(R_i^{'})^{\textit{best}} \leftarrow R_i$

    $\textit{MinCost} \leftarrow \infty$

        \uIf {$\left|Y_{R_i}\right| = 0$}
        {
       
            $ (R_i^{'})^{\textit{best}} , \textit{MinCost}  \leftarrow \textbf{Adjust}({R_i}, 0, 0, T_j, \textit{ratio}, L^{\textit{depot}}, D^{\textit{dwell}}, V)$
        }
        \ElseIf{$dw_j^{\textit{end}} + < R_i^{\textit{start}} + D^{\textit{maxroute}}$} {

            $\textit{index} \leftarrow 0$
            
            $Placements^{\textit{pickup}} \leftarrow \textbf{GetPlacements}({R_i}, L^{\textit{pickup}}_j, w_j, \textit{index}, D^{\textit{dwell}})$
            
            $Placements^{\textit{dropoff}} \leftarrow \emptyset$
            
            \If {$\left|Placements^{\textit{pickup}}\right| > 0$}
            {
            
               $\textit{index}  \leftarrow \min(Placements^{\textit{pickup}}) - 1$
            
                $Placements^{\textit{dropoff}} \leftarrow \textbf{GetPlacements}(R_i, L^{\textit{dropoff}}_j,  dw_j, \textit{index}, D^{\textit{dwell}})$
                
            }
            
            \If {$\left|Placements^{\textit{pickup}}\right| \times \left|Placements^{\textit{dropoff}}\right| > 0$}
            {
            
            $MinCost \leftarrow \infty$
            
                \For {$p_{idx} \in Placements^{\textit{pickup}}$}
                {
                    \For {$d_{idx} \in Placements^{\textit{dropoff}}$}
                    {
                     
                     \If {$p_{idx} < d_{idx}$}
                     {

                        $R_i^{'}, \textit{SolCost} \leftarrow \textbf{Adjust}({R_i}, p_{idx}, d_{idx}, T_j, \textit{ratio}, L^{\textit{depot}}, \newline D^{\textit{dwell}}, V)$

                        \If {$\textit{SolCost}\neq \infty$}
                        {
      
                          $threshold \leftarrow \textbf{GThreshold}(\textit{ratio}, R_i^{'})$
                            
                            \If {$\textit{SolCost} < threshold \land \textit{SolCost} < MinCost$}
                            {
                                
                                $MinCost \leftarrow \textit{SolCost}$
                               
                                   $(R_i^{'})^{\textit{best}} \leftarrow R_i^{'}$

                            }
                        }
    
                    }
                     
                     }

                }
            }
        }
\KwResult{ $(R_i^{'})^{\textit{best}}, \textit{MinCost} $}
\end{algorithm}

\cref{algo:feasible} checks whether each request $T_j \in \RequestsVector$ can be inserted into the given route $R_i$ based on existing nodes  ($Y_{R_i}$) in the route $R_i$. This algorithm considers two cases; first, when the route is empty (i.e., no nodes exist for the route), this case is trivial. The algorithm adds the request by just appending the pickup node $\langle L^{\textit{pickup}}_j, w_j \rangle$ and drop-off node $\langle L^{\textit{dropoff}}_j, dw_j \rangle$ without checking any time constraints and capacity constraints. Next, when the route is not empty, the algorithm first obtains all the possible placements for the pickup node $\langle L^{\textit{pickup}}_j, w_j \rangle$. If it can obtain at least one placement, then the algorithm tries to obtain all the possible placements for drop-off node $\langle L^{\textit{dropoff}}_j, dw_j \rangle$. If it has feasible placements for both the pickup  and drop-off nodes, then the algorithm checks the feasibility of each pair of placement values using \textbf{Adjust}. If the assignment is feasible (i.e., the $\textit{SolCost} \neq \infty$), then the algorithm computes threshold value using the function \textbf{GThreshold}. We can formally express the function \textbf{GThreshold} as follows:
\begin{align}
    &GThreshold(\textit{ratio}, R_i') =\nonumber \\ &p_{const} + p_{ratio-thres} \cdot \textit{ratio} + p_{length} \cdot ((R_i')^{\text{end}} - (R_i')^{\textit{start}})
    \label{eqn:gthreshold}
\end{align}

If the $\textit{SolCost}$ is less than the threshold value and less than the current minimum cost, then update the current minimum cost with the $\textit{SolCost}$ and update the best-updated route ($(R_i')^{\textit{best}}$), and update the extra time and wait time corresponding to the minimum cost. At the end of this iterative process, the algorithm will return whether a feasible assignment exists for the request ($T_j$), the best-updated route ($(R_i')^{\textit{best}}$), and the cost $\textit{MinCost}$ corresponding  corresponding to the best-updated route. The running time of this algorithm is $\mathcal{O}(\left|\RequestsVector\right|^3)$.

\begin{algorithm}[!h]
 \caption{$\textbf{Adjust}(R_i, p_{idx}, d_{idx}, T_j, \textit{ratio},  L^{\textit{depot}}, \newline D^{{\textit{dwell}}},  D^{\textit{maxroute}}, V)$}
 \label{algo:adjust}

$Y_{R_i^{'}} \leftarrow \textbf{AdjustNodes}(R_i, p_{idx}, d_{idx})$
                                     
$success_d, D_{R_i^{'}} \leftarrow \textbf{AdjustTimes}(Y_{R_i^{'}}$)
    
$success_o, O_{R_i^{'}} \leftarrow \textbf{AdjustCapacities}(Y_{R_i^{'}}, V)$

$\textit{SolCost} \leftarrow \infty$
                    
\If {$success_d \land success_o$}
{

$l^{\textit{first}}, w^{\textit{first}} \leftarrow Y_{R_i^{'}}^{\textit{first}}$

$l^{\textit{last}}, w^{\textit{last}} \leftarrow Y_{R_i^{'}}^{\textit{last}}$

$D^{\textit{source}} \leftarrow D^{\textit{travel}}_{L^{\textit{depot}}, l^{\textit{first}}}$

$D^{\textit{sink}} \leftarrow D^{\textit{travel}}_{l^{\textit{last}}, L^{\textit{depot}}}$

$D^{\textit{newduration}} \leftarrow D_{R_i^{'}}^{\textit{last}} - D_{R_i^{'}}^{\textit{first}} + D^{\textit{source}} + D^{\textit{sink}} + D^{{\textit{dwell}}}$

\If {$D^{\textit{newduration}} \leq D^{\textit{maxroute}}$}
{

$D^{\textit{extra}} \leftarrow D^{\textit{newduration}} - {R_i}^{\textit{end}} - {R_i}^{{\textit{start}}}$

$D^{\textit{wait}} \leftarrow 0 $

\If {$Y_{R_i^{'}} = Y_R + \{(L^{\textit{pickup}}_j,w_j), (L^{\textit{dropoff}}_j,dw_j) \}$}
{

\If {$|Y_R| > 0 $}
{
    $l_{\textit{prev}}, w_{\textit{prev}} \leftarrow Y_R^{\textit{last}}$

    $D^{\textit{wait}} \leftarrow \max(0, w_{\textit{prev}}^{{\textit{start}}} - D^{\textit{travel}}_{l_{\textit{prev}},L^{\textit{pickup}}_j }  - D^{{\textit{dwell}}})$
}

}

$\textit{SolCost} \leftarrow \textbf{GCost}(\textit{ratio}, D^{\textit{extra}}, D^{\textit{wait}})$
    
}

}

\KwResult{${R_i^{'}}, \textit{SolCost}$}
\end{algorithm}
\cref{algo:adjust} checks whether the assignment of a new request to a given pickup placement ($p_{\textit{idx}}$) and dropoff placement  ($d_{\textit{idx}}$) violates any travel-time, capacity, or route length constraints (such as maximum route duration). 
If the assignment does not violate any constraints, then  the algorithm computes the increase in the route duration $(D^\textit{extra})$ for route $R_i$, as well as the additional wait time $D^\textit{wait}$ before serving the request $T_j$ in the current route $R_i$ if the algorithm assign request $T_j$ to the route $R_i$ by placing the pickup node at the pickup placement value $(p_{\textit{idx}})$ and placing the drop-off node at the drop-off placement value  $(d_{\textit{idx}})$.   
Finally, the algorithm computes the cost based on the function \textbf{GCost} (see \cref{eqn:gcost}) with increase in the route duration $(D^\textit{extra})$, additional wait time $(D^\textit{wait})$, and the ratio of requests assigned so far ($\textit{ratio}$). We can formally express the function \textbf{GCost} as follows,
\begin{align}
    &GCost(\textit{ratio}, D^{\textit{extra}}, D^{\textit{wait}}) =\nonumber \\ & D^{\textit{extra}} + (p_{wait} + p_{ratio-cost} \cdot \textit{ratio}) \cdot  D^{\textit{wait}}
    \label{eqn:gcost}
\end{align}

The algorithm returns the updated route ($(R_i')^{\textit{best}}$) and the corresponding cost $\textit{SolCost}$. Algorithms \textbf{AdjustTimes} and \textbf{AdjustCapacities} have a time complexity of  $\mathcal{O}(\left|\RequestsVector\right|)$, whereas  algorithm \textbf{AdjustNodes} has a time complexity of $\mathcal{O}(1)$. Hence, the computation time of this algorithm is $\mathcal{O}(\left|\RequestsVector\right|)$.

\begin{algorithm}[!ht]
 \caption{$\textbf{BestAssignment}(R_i, \RequestsVector, \textit{ratio}, L^{\textit{depot}}, D^{\textit{dwell}},\newline D^{\textit{maxroute}}, V)$}
 \label{algo:bestassignment}

MinCost $\leftarrow$ $\infty$

$(R_i^{'})^{\textit{best}} \leftarrow R_i$
    
$(T_j)^{best} \leftarrow None$

\For {$T_j \in \RequestsVector$}
{

    $R_i^{'}, \textit{SolCost} \leftarrow \textbf{Feasible}(R_i, T_j, \textit{ratio}, L^{\textit{depot}}, D^{\textit{dwell}}, D^{\textit{maxroute}}, V)$
    
    \If {$\textit{SolCost} \neq \infty$}
    {

        \If {$\textit{SolCost} < \textit{MinCost}$}
        {
        
            $\textit{MinCost} \leftarrow \textit{SolCost}$
            
            $(R_i^{'})^{\textit{best}} \leftarrow R_i$
                
            $(T_j)^{best} \leftarrow T_j$

        }

    }

}

\KwResult{$(R_i^{'})^{\textit{best}}, (T_j)^{\textit{best}}$}
\end{algorithm}

\cref{algo:bestassignment} finds the request that has the  lowest weighted cost when the algorithm inserts the request into the given route $R_i$ based on existing nodes ($Y_{R_i}$) for the route $R_i$ and remaining requests ($\RequestsVector$). The algorithm iterates over each request $T_j \in \RequestsVector$ and first checks whether the algorithm can add  request $T_j$ to  route $R_i$. If it can, then the algorithm computes the weighted cost. At the end of this process, the algorithm returns the request ($(T_j)^{\textit{best}}$) with the lowest weighted cost and the updated best route ($(R_i')^{\textit{best}}$). The running time of this algorithm is $\mathcal{O}(\left|\RequestsVector\right|^4)$.

\begin{algorithm}[!ht]
 \caption{$\textbf{Greedy}(\RequestsVector,  L^{depot}, D^{\textit{dwell}}, D^{\textit{maxroute}}, V)$}
 \label{algo:greedy}

$i \leftarrow 0$

$\RunsVector \leftarrow \emptyset$

$t^{\textit{size}} \leftarrow \left|\RequestsVector\right|$

\While {$\left|\RequestsVector\right| >$ 0}
{

    $R_i \leftarrow \textbf{CreateEmptyRun}(i)$
    
    \While {$\left|\RequestsVector\right| >$ 0}
    {
        
        $\textit{ratio} \leftarrow \frac{\left|\RequestsVector\right|}{t^{\textit{size}}}$
        
        $R_i^{'}, T_j \leftarrow \textbf{BestAssignment}(R_i,\RequestsVector, \textit{ratio}, L^{\textit{depot}}, \newline D^{\textit{dwell}}, D^{\textit{maxrun}}, V
        )$
        
        \uIf {$T_j \neq None$}
        {

            $R_i \leftarrow R_i^{'}$
            
            $\RequestsVector \leftarrow \RequestsVector \setminus \{ T_j \}$
            
        }
        \Else
        {
                    break
        }
    }
    
    $\RunsVector \leftarrow \RunsVector \cup \{R_i\}$

    $i \leftarrow i + 1$
            
}
\KwResult{$\RunsVector$}
\end{algorithm}

\cref{algo:greedy} assigns all the available requests $\RequestsVector$. The algorithm follows an iterative process, where in each iteration, the algorithm generates a new route $R_i$ and tries to add requests one-by-one from $\RequestsVector$.  When no more feasible requests are available for the route $R_i$,  but there are more requests remaining to be assigned, then the algorithm generates the next route and repeats the process. This iterative process  terminates when the algorithm has assigned all the requests to routes. The running time of this algorithm is $\mathcal{O}(\left|\RequestsVector\right|^5)$.

\clearpage
\section{Supplementary Numerical Results}
\label{app:numerical}
\pgfplotstableread[col sep=comma,]{data/comparison_2/comparison_sim_anneal_anytime_120.csv}\ourctwo
\pgfplotstableread[col sep=comma,]{data/comparison_2/comparison_sim_anneal_anytime_180.csv}\ourcthree
\pgfplotstableread[col sep=comma,]{data/comparison_2/comparison_sim_anneal_anytime_240.csv}\ourcfour

\pgfplotstableread[col sep=comma,]{data/comparison_6/comparison_routing_anytime_120.csv}\routingctwo
\pgfplotstableread[col sep=comma,]{data/comparison_6/comparison_routing_anytime_180.csv}\routingcthree
\pgfplotstableread[col sep=comma,]{data/comparison_6/comparison_routing_anytime_240.csv}\routingcfour

\pgfplotstableread[col sep=comma,]{data/comparison_2/comparison_vroom_anytime_120.csv}\vroomctwo
\pgfplotstableread[col sep=comma,]{data/comparison_2/comparison_vroom_anytime_180.csv}\vroomcthree
\pgfplotstableread[col sep=comma,]{data/comparison_2/comparison_vroom_anytime_240.csv}\vroomcfour

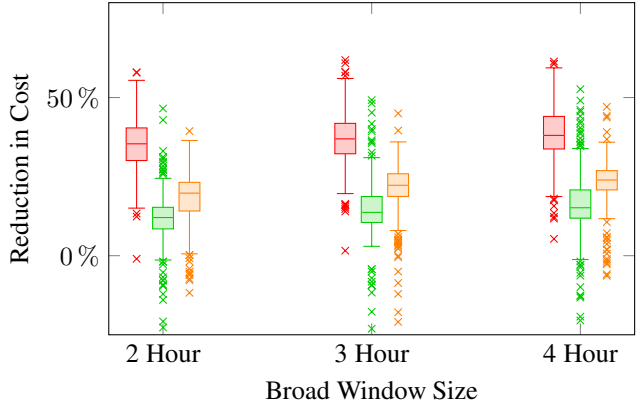
\begin{figure}
\begin{tikzpicture}
\begin{axis}[
      boxplot/draw direction=y,
      xtick={1,2,3},
      xticklabels={{2 Hour}, {3 Hour}, {4 Hour}},
      width=\columnwidth,
      height = 6cm,
      bugsResolvedStyle/.style={},
      ylabel={Reduction in Cost},
      yticklabel=\pgfmathprintnumber{\tick}\,$\%$,
      xlabel={Broad Window Size},
      ymin=-25,
      ymax=80,
    ]
\addplot+[boxplot={box extend=0.10, draw position=1},ColorOur, solid, fill=ColorOur!20, mark=x] table [col sep=comma, y=diff_sim_anneal_anytime] {\ourctwo};
\addplot+[boxplot={box extend=0.10, draw position=2}, ColorOur, solid, fill=ColorOur!20, mark=x] table [col sep=comma, y=diff_sim_anneal_anytime] {\ourcthree};
\addplot+[boxplot={box extend=0.10, draw position=3}, ColorOur, solid, fill=ColorOur!20, mark=x] table [col sep=comma, y=diff_sim_anneal_anytime] {\ourcfour};

\addplot+[boxplot={box extend=0.10, draw position=1},ColorVroom, solid, lshift, fill=ColorVroom!20, mark=x] table [col sep=comma, y=diff_vroom_anytime] {\vroomctwo};
\addplot+[boxplot={box extend=0.10, draw position=2}, ColorVroom, solid, lshift, fill=ColorVroom!20, mark=x] table [col sep=comma, y=diff_vroom_anytime] {\vroomcthree};
\addplot+[boxplot={box extend=0.10, draw position=3}, ColorVroom, solid,lshift, fill=ColorVroom!20, mark=x] table [col sep=comma, y=diff_vroom_anytime] {\vroomcfour};

\addplot+[boxplot={box extend=0.10, draw position=1},ColorGoogle, solid, rshift, fill=ColorGoogle!20, mark=x] table [col sep=comma, y=diff_routing_anytime] {\routingctwo};
\addplot+[boxplot={box extend=0.10, draw position=2}, ColorGoogle, solid, rshift, fill=ColorGoogle!20, mark=x] table [col sep=comma, y=diff_routing_anytime] {\routingcthree};
\addplot+[boxplot={box extend=0.10, draw position=3}, ColorGoogle, solid,rshift, fill=ColorGoogle!20, mark=x] table [col sep=comma, y=diff_routing_anytime] {\routingcfour};

\end{axis}
\end{tikzpicture}
\caption{Reduction in total cost due to using our approach ($\mu^*$ and $\alpha^\textit{SimAnn+Greedy}$) for selecting tight pickup windows, compared to using na\"ive window selection, with three different offline VRP solvers: VROOM  (\textcolor{ColorLegendVroom}{$\blacksquare$}), Google OR-Tools  (\textcolor{ColorLegendGoogle}{$\blacksquare$}), and our greedy and simulated annealing algorithms as offline VRP solvers  (\textcolor{ColorLegendOur}{$\blacksquare$}).
}
\label{fig:ourbest_vs_static_for_all_vrp}
\end{figure}

\subsubsection{Hyperparameter Search}

We first perform a grid search to find optimal hyperparameters for the greedy algorithm. Based on the results of this search, we set values for the wait-time $p_{wait} = 0.1$ and fraction of requests served   $p_{ratio-cost} = 0.1$ parameters, which are used by the algorithm  to estimate the cost of assigning a request to a route (see \cref{algo:adjust}). We also set values for the length of the route $p_{length} = 10$, the fraction of requests served $p_{ratio-thres} = 0.1$, and the constant value of the threshold $p_{const} = 0.1$ parameters, which restrict the assignment of requests to routes (see \cref{algo:feasible}).

We next perform a grid search to find optimal hyperparameters for the simulated annealing algorithm. Based on the results of this search, we set the altering rate to $p_{alter} = 0.4$ (see \cref{algo:randomneighbor}) and the initial and final probabilities to $p_{start} = 0.9$ and $p_{end} = 0.5$, respectively (see \cref{algo:simulatedannealing}).

\subsubsection{Proposed Approach vs. Na\"ive Pickup Window}


Here, we present additional numerical results showing the advantage of using our proposed approach to select tight pickup windows.
\cref{fig:ourbest_vs_static_for_all_vrp} shows the reduction in the total cost of vehicle routes due to using our approach (policy $\mu^*$ supported by anytime algorithm $\alpha^\textit{SimAnn+Greedy}$) to select tight pickup windows, compared to na\"ively selected pickup windows.
We consider three offline VRP solvers to compare our windows with the na\"ive ones: VROOM, Google OR-Tools, and our greedy and simulated annealing algorithms applied as offline VRP solvers.
In each case, we evaluate the algorithms on 160 days of paratransit data, and plot the distributions.
We observe that when we use our algorithms as offline VRP solvers, there is a significant, 14-18\% reduction in cost, which is consistent with \cref{fig:ourrtat_vs_ourrt_sa_vs_ourg_sa}.
Again, we see that longer broad pickup windows lead to more pronounced reduction since they provide more flexibility for optimizing the tight pickup windows.

On the other hand, we see a less significant reduction in cost when using VROOM and Google OR-Tools: 4-6\% and 1-2\% reduction, respectively.
We hypothesise that this is explained by the fact that we train our policy $\mu^*$ with our algorithms ($\alpha^\textit{SimAnn+Greedy}$ as the anytime and $\alpha^\textit{Greedy}$ as the shaped cost $\tilde{c}_i$ estimator); hence, our policy $\mu^*$ learns to work well with our VRP solvers.
Since VROOM, Google OR-Tools, and our algorithms are all heuristic, the tight windows selected by our policy $\mu^*$ may not work well with VROOM and Google OR-Tools.
We can address this by training a policy $\mu$ with VROOM and with Google OR-Tools, so that the policy $\mu$ will select tight windows that fit these VRP solvers.
We will investigate this in future work.


\subsubsection{Training Process}

\pgfplotstableread[col sep=comma,]{data/model/models_nn.csv}\modeleval

\begin{figure}
\begin{tikzpicture}
\begin{axis}[
      width=\columnwidth,
      height = 5.75cm,
      bugsResolvedStyle/.style={},
      ylabel={Total Cost},
      xlabel={Training Episodes},
    ]
\addplot table [x=model_idx, y=objective, col sep=comma] {data/model/models_nn.csv};
\end{axis}
\end{tikzpicture}
\caption{Evolution of the performance of the policy $\mu$, measured as the average total cost of the resulting VRPs, throughout the training process.
Each datapoint is based on 5 distinct policies (trained for the given number of episodes), which are evaluated on 5 different problem instances each. 
}
\label{fig:model_analysis}
\end{figure}
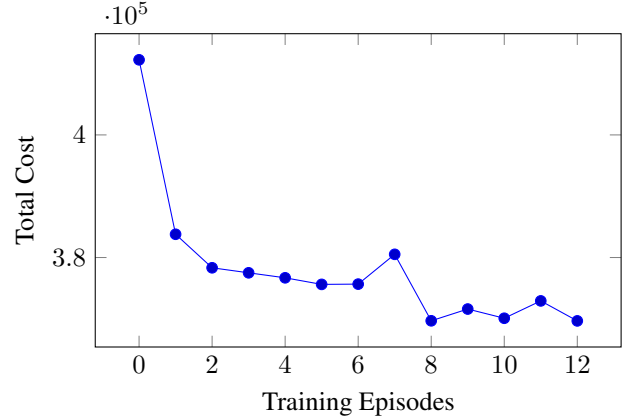

\pgfplotstableread[col sep=comma,]{data/model/models_nn_full.csv}\modelevalfull

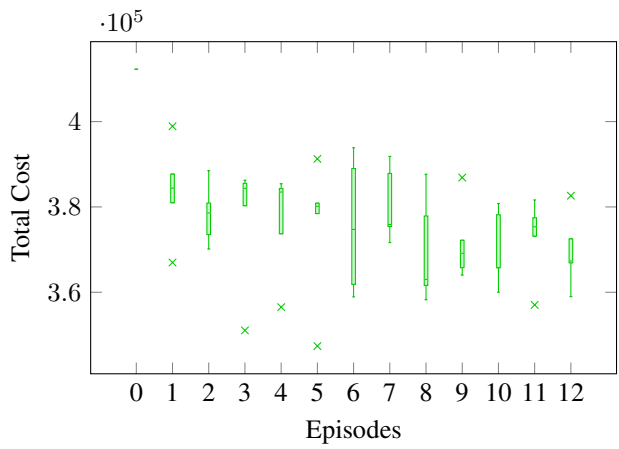
\begin{figure}
\begin{tikzpicture}
\begin{axis}[
      boxplot/draw direction=y,
      xtick={1,2,3,4,5,6,7,8,9,10,11,12,13},
      xticklabels={{0}, {1}, {2}, {3}, {4}, {5}, 
      {6}, {7}, {8}, {9}, {10}, {11}, 
      {12}},
      width=\columnwidth,
      height = 6cm,
      bugsResolvedStyle/.style={},
      ylabel={Total Cost},
      xlabel={Episodes},
    ]
\addplot+[boxplot={box extend=0.10, draw position=1},ColorOur, solid, fill=ColorOur!20, mark=x] table [col sep=comma, y=model_0] {\modelevalfull};
\addplot+[boxplot={box extend=0.10, draw position=2}, ColorOur, solid, fill=ColorOur!20, mark=x] table [col sep=comma, y=model_1] {\modelevalfull};
\addplot+[boxplot={box extend=0.10, draw position=3}, ColorOur, solid, fill=ColorOur!20, mark=x] table [col sep=comma, y=model_2] {\modelevalfull};
\addplot+[boxplot={box extend=0.10, draw position=4}, ColorOur, solid, fill=ColorOur!20, mark=x] table [col sep=comma, y=model_3] {\modelevalfull};
\addplot+[boxplot={box extend=0.10, draw position=5}, ColorOur, solid, fill=ColorOur!20, mark=x] table [col sep=comma, y=model_4] {\modelevalfull};
\addplot+[boxplot={box extend=0.10, draw position=6}, ColorOur, solid, fill=ColorOur!20, mark=x] table [col sep=comma, y=model_5] {\modelevalfull};
\addplot+[boxplot={box extend=0.10, draw position=7}, ColorOur, solid, fill=ColorOur!20, mark=x] table [col sep=comma, y=model_6] {\modelevalfull};
\addplot+[boxplot={box extend=0.10, draw position=8}, ColorOur, solid, fill=ColorOur!20, mark=x] table [col sep=comma, y=model_7] {\modelevalfull};
\addplot+[boxplot={box extend=0.10, draw position=9}, ColorOur, solid, fill=ColorOur!20, mark=x] table [col sep=comma, y=model_8] {\modelevalfull};
\addplot+[boxplot={box extend=0.10, draw position=10}, ColorOur, solid, fill=ColorOur!20, mark=x] table [col sep=comma, y=model_9] {\modelevalfull};
\addplot+[boxplot={box extend=0.10, draw position=11}, ColorOur, solid, fill=ColorOur!20, mark=x] table [col sep=comma, y=model_10] {\modelevalfull};
\addplot+[boxplot={box extend=0.10, draw position=12}, ColorOur, solid, fill=ColorOur!20, mark=x] table [col sep=comma, y=model_11] {\modelevalfull};
\addplot+[boxplot={box extend=0.10, draw position=13}, ColorOur, solid, fill=ColorOur!20, mark=x] table [col sep=comma, y=model_12] {\modelevalfull};
\end{axis}
\end{tikzpicture}
\caption{Evolution of the performance of the policy $\mu$, measured as the average total cost of the resulting VRPs, throughout the training process.
Each box plot is based on 5 distinct policies (trained for the given number of episodes), which are evaluated on 5 different problem instances each.
}
\label{fig:model_analysis_updated}
\end{figure}

Finally, we present numerical results on the reinforcement-learning process to show how the performance of the policy $\mu$ improves as it is trained on more and more episodes.
Since the reinforcement-learning process is non-deterministic (as the neural-network representation of~$Q$ is initialized at random, and random actions are chosen occasionally to ensure exploration), we provide robust results by initializing 5 different neural-network instances and then training these instances independently of each other over a number of episodes.
After each episode, we evaluate each policy (i.e., each neural-network instance) on 5 problem instances (i.e., 5 days of paratransit data).
\cref{fig:model_analysis} shows the average performance of these 5 policies over the 5 problem instances, for training episodes 0 (i.e., untrained networks) to 12.

We observe that trained policies (more than 0 episodes) have a significant advantage over untrained policies (0 episodes), which select tight windows at random (within the broad windows) since the neural networks are initialized with random weights.
Further, we see that the performance curve starts to ``flatten out'' around 8-12 episodes, which suggests that we may not need many more episodes to find optimal policies.


\clearpage
\section{Dataset Statistics}
\label{app:data_analysis}
In this section, we study how the spatial and temporal distribution of the trips varies based on the booking time. Accordingly, we categorize the requests into the requests made in the morning (9am - 12pm), afternoon (12pm - 3pm), and evening (3pm - 5pm). Then, we look into the spatial and temporal distribution of the requests based on these three categories.

\pgfplotstableread[col sep=comma,header=true]{data_analysis/temporal_full.csv}\temporal

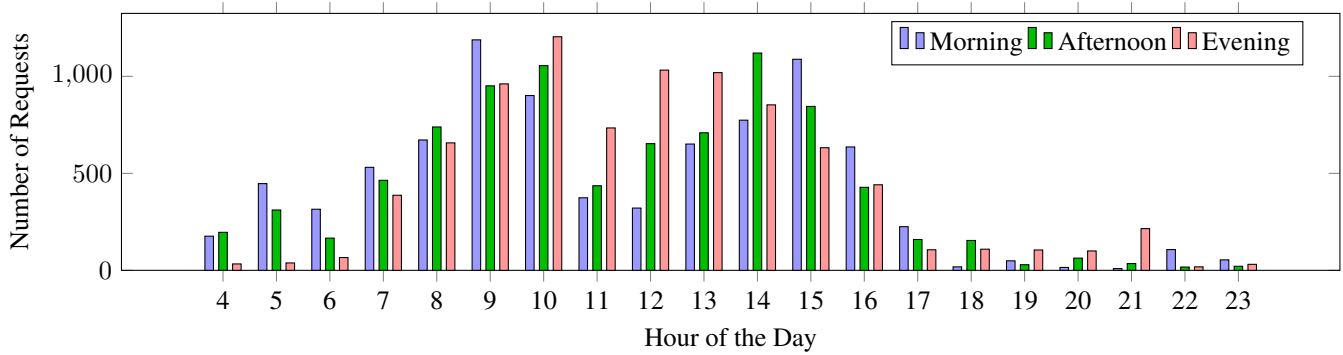
\begin{figure*}
\centering
\begin{tikzpicture}
\begin{axis}[
    legend style={at={(0.485, 0.99)},anchor=north east},
    ybar, ymin=0,
    ylabel style={align=center},
    ylabel=Number of Requests,
    xlabel=Hour of the Day,
    xtick=data,
    width=\linewidth,
    height = 5cm, 
    legend pos=north east,
    legend columns=3,
    xticklabels from table={\temporal}{hour}
    ]
    \addplot [BlueBars] table[ x expr=\coordindex, y expr=\thisrow{morning_count}] {\temporal};
    \addlegendentry{Morning}
    \addplot [GreenBars] table[ x expr=\coordindex, y expr=\thisrow{afternoon_count}] {\temporal};
    \addlegendentry{Afternoon}
    \addplot [RedBars] table[ x expr=\coordindex, y expr=\thisrow{evening_count}] {\temporal};
    \addlegendentry{Evening}
\end{axis}
\end{tikzpicture}
\caption{Distribution of the pickup time of the requests based on the booking time}
\label{fig:temporal}
\end{figure*}

The \cref{fig:temporal} shows the distribution of the pickup time of requests based on the time of booking. In the figure, we observe that customers are more likely to book requests for the next day early morning, either in the morning or afternoon of the previous day. Similarly, the customers are more likely to book requests for the next day late evening in the evening of the booking day. 

\pgfplotstableread[col sep=comma,header=true]{data_analysis/spatial_full.csv}\spatial

\begin{figure*}
\centering
\begin{tikzpicture}
\begin{axis}[
    legend style={at={(0.485, 0.99)},anchor=north east},
    ybar, ymin=0,
    ylabel style={align=center},
    ylabel=Number of Requests,
    xlabel=Anonymized ZIP codes,
    xtick=data,
    width=\linewidth,
    height = 5cm, 
    legend pos=north east,
    legend columns=3,
    xticklabels={1,2,3,4,5,6,7,8,9,10,11,12,13,14,15,16,17,18,19,20,21,22},
    ]
    \addplot [BlueBars] table[ x expr=\coordindex, y expr=\thisrow{morning_count}] {\spatial};
    \addlegendentry{Morning}
    \addplot [GreenBars] table[ x expr=\coordindex, y expr=\thisrow{afternoon_count}] {\spatial};
    \addlegendentry{Afternoon}
    \addplot [RedBars] table[ x expr=\coordindex, y expr=\thisrow{evening_count}] {\spatial};
    \addlegendentry{Evening}
\end{axis}
\end{tikzpicture}
\caption{Distribution of the pickup location of the requests based on the booking time}
\label{fig:spatial}
\end{figure*}
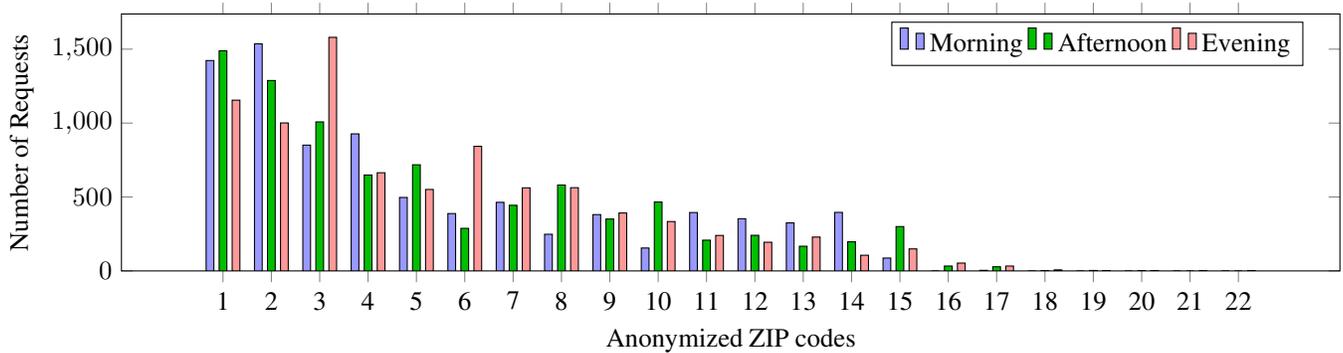


The \cref{fig:spatial} shows the distribution of the pickup location (ZIP code) of the requests based on the time of the booking. We shifted the coordinates to anonymize the data. The distribution of pickup locations does not vary much between the requests made in the morning and afternoon. Moreover, the distribution of pickup locations varies considerably between requests made in the evening and the rest of the day.

These observations indicate that there is a considerable difference between the pickup location and pickup time, between the requests booked in the morning, afternoon, and evening. 

\clearpage
\section{Extended Discussion of Related Work}
\label{app:related}

We can divide prior work on vehicle routing problems into two main categories: offline (or day-ahead) VRPs and online (or real-time) VRPs.
We presented a discussion of relate work focusing on this distinction in \cref{sec:related}.
Here, we provide a further discussion of related work focusing on the objective formulation of VRPs as well as the approaches used to solve dial-a-ride problems.

\subsection*{VRP Objectives}

Vehicle routing problems can also be categorized based on their objectives, that is, based on how they quantify the cost of a VRP solution. \citeSubject{agatz2011dynamic}, \citeSubject{turmo2018evaluating}, \citeSubject{alonso2017demand}, \citeSubject{ota2016stars}, \citeSubject{simonetto2019real}, and \citeSubject{wen2018transit} consider minimizing the total miles traveled by the vehicles.
Meanwhile, \citeSubject{turmo2018evaluating}, \citeSubject{gupta2010improving}, and \citeSubject{simonetto2019real} consider minimizing the total hours traveled by the vehicles, similar to our VRP formulation.
In contrast, \citeSubject{alonso2017demand}, \citeSubject{turmo2018evaluating}, and \citeSubject{wen2018transit} focus on improving the quality of VRP solutions from the perspective of passengers by minimizing the passengers' total waiting time. Similarly, \citeSubject{salazar2018interaction} consider reducing the total passenger travel time. In our VRP formulation, we consider minimizing the total travel time of the vehicles and the number of vehicle routes.





\subsection*{Dial-a-Ride Problem}
Previous research efforts use approaches such as tabu search \cite{mo2018mass,berbeglia2012hybrid}, greedy approach \cite{qian2017optimal,alonso2017demand}, and exact methods \cite{liu2015branch,parragh2015dial,gschwind2015effective} to solve the scheduling in dial-a-ride problem and paratransit operations.

\subsection*{Pickup and Delivery Problem}
In earlier research efforts, researchers used adaptive neighbourhood search \cite{ropke2006adaptive}, and exact-methods \cite{ropke2007models,ropke2009branch,qu2015branch} to solve the pickup and delivery problem.

\fi


\end{document}